\documentclass{article}
\usepackage{ijcai19}

\usepackage{times}
\usepackage{soul}
\usepackage{url}
\usepackage[hidelinks]{hyperref}
\usepackage[utf8]{inputenc}
\usepackage[small]{caption}
\usepackage{graphicx}
\usepackage{amsmath}
\usepackage{booktabs}
\usepackage{float}
\usepackage[caption = false]{subfig}
\usepackage{amsfonts}
\usepackage{amsmath}

\newcommand{\norm}[1]{\left\lVert#1\right\rVert}

\title{The Effect of Planning Shape on Dyna-style Planning\\ in
  High-dimensional State Spaces}

\author{
G. Zacharias Holland$^1$
\and
Erin J. Talvitie$^2$\and
Michael Bowling$^1$
\affiliations
$^1$University of Alberta\\
$^2$Franklin \& Marshall College\\
\emails
gholland@ualberta.ca,
erin.talvitie@fandm.edu,
mbowling@ualberta.ca
}

\begin{document}

\maketitle

\begin{abstract}  
  Dyna is a fundamental approach to model-based reinforcement learning
  (MBRL) that interleaves planning, acting, and learning in an online
  setting. In the most typical application of Dyna, the dynamics model
  is used to generate one-step transitions from selected start states
  from the agent's history, which are used to update the agent's value
  function or policy as if they were real experiences. In this work,
  one-step Dyna was applied to several games from the Arcade Learning
  Environment (ALE). We found that the model-based updates offered
  surprisingly little benefit over simply performing more updates with
  the agent's existing experience, even when using a perfect model. We
  hypothesize that to get the most from planning, the model must be
  used to generate unfamiliar experience. To test this, we
  experimented with the ``shape'' of planning in multiple different
  concrete instantiations of Dyna, performing fewer, longer rollouts,
  rather than many short rollouts. We found that planning shape has a
  profound impact on the efficacy of Dyna for both perfect and learned
  models. In addition to these findings regarding Dyna in general, our results
  represent, to our knowledge, the first time that a learned dynamics
  model has been successfully used for planning in the ALE, suggesting
  that Dyna may be a viable approach to MBRL in the ALE and other
  high-dimensional problems.
\end{abstract}

\section{Introduction}
Dyna \cite{sutton1990dynaq} is a general architecture for
reinforcement learning (RL) agents that flexibly combines aspects of
both model-free and model-based RL. In Dyna, the agent learns a
predictive model of its environment, which is used to generate
simulated experience alongside the agent's real experience. Both types
of experience are used in the same way: to update the agent's value
function and/or policy.  Planning in this way is referred to as
\textit{Dyna-style planning}.

Dyna-style planning has many appealing properties that make it a
potentially powerful approach in large-scale problems. Because the
model indirectly influences the agent's behavior via the value
function or policy, the computationally expensive process of planning
can be asynchronous with the agent's decision-making loop,
retaining a model-free agent's ability to operate on a fine-grained
timescale. Furthermore, Dyna permits flexible control over how much
computational effort is devoted to planning and how that planning
effort is distributed, potentially focusing on updates that will have
the most impact, as in Prioritized Sweeping
\cite{moore1993prioritized}. That said, there have been relatively few
examples of the application of Dyna-style planning in large-scale
problems with high-dimensional observations, such as images.

In this work we study the behavior of Dyna in the Arcade Learning
Environment (ALE) \cite{bellemare13arcade}, where agents learn to play
games from the Atari 2600 system with raw images as input. Though many
successful methods exist for playing games from the ALE and other
image based domains
(e.g. \cite{mnih2015human,mnih2016asynchronous,hessel2018rainbow}),
the majority of these approaches can be considered
\textit{model-free}, in that they learn and improve a policy without
using a predictive model of the environment. Model-based methods have
the potential to offer dramatic sample complexity benefits and are
thus of significant interest. However, despite the introduction of
increasingly effective approaches for learning predictive models in
Atari games
\cite{bellemare2013bayesian,bellemare2014skip,oh2015action}, the
application of model-based methods to the ALE is still an open
problem, with many challenges to overcome \cite{machado17arcade}.

Though some recent approaches to image-based problems incorporate
model-like components
(e.g. \cite{tamar2016value,oh2017value,weber2017imagination}), these
methods arguably discard many of the anticipated benefits of
model-learning by making the model essentially a part of the
parameterization of the value function or policy. Part of the allure
of model-based RL is that a dynamics model could be learned even when
the agent has little guidance on how to behave and that the model can
be used as a substitute for gathering expensive experience in the real
world. In this work we focus on learning and directly planning with a
dynamics model.

In the original Dyna-Q algorithm \cite{sutton1990dynaq}, planning was
performed by selecting several states from the agent's recent
experience and using the model to roll forward one step from each of
those states and thus generate value function updates. This remains
the most common form of Dyna-style planning. In our experiments in the
ALE we found that using the model in this way provided surprisingly
little benefit, even with a perfect model --- similar sample
efficiency gains could be obtained simply by performing more
model-free updates on the data the agent had already gathered. We
note, however, that Dyna-style planning can take on a variety of
shapes. For a fixed budget of model interactions, one might generate
many one-step rollouts or fewer multi-step rollouts. Empirically we
find that planning with longer rollouts yields dramatic improvement
when using a perfect model. We also find that planning shape impacts
performance when the model is learned (and is therefore imperfect),
though model flaws can make rollouts unreliable.

The primary contribution of this paper is an empirical exploration of
the impact of planning shape on Dyna-style planning. The results offer
guidance to future practitioners wishing to apply Dyna. In particular,
we find that:
\begin{enumerate}
\item For Dyna to take full advantage of a model, it must use the
  model to generate unfamiliar experience (in essence to replace
  exploration in the real environment).
\item Performing longer rollouts in the model seems to be a simple,
  effective way to generate unfamiliar experience.
\item Planning shape seems to be an important consideration even when
  the model is imperfect.
\item Even if there were dramatic improvements in model-learning,
  where highly accurate models could be learned very quickly, there
  would still be little benefit from planning with one-step
  rollouts. Longer rollouts are necessary to allow planning to exploit
  more accurate models.
\end{enumerate}

Furthermore, though this paper is primarily focused on the above
scientific observations and not on generating new state of the art
results in the ALE, the experimental results demonstrate for the first
time (as far as we are aware) a sample-complexity benefit from
learning a dynamics model in some games. This benefit is at its
greatest with proper choice of planning shape. Model-based
reinforcement learning in the ALE has proven to be such a challenge
that even a modest benefit from a learned model is a significant and
unprecedented development \cite{machado17arcade}. This finding may
point the way toward yet more effective model-based approaches in this
domain and establishes Dyna as a viable planning method
in large-scale, high-dimensional domains.

\section{The  Ineffectiveness of One-step Rollouts}
Planning in the Dyna architecture is accomplished by using a model to
make predictions of future states based on a start state and action.
When the state is high dimensional, like an image, it is not clear how
to generate reasonable start states. One solution is to sample the
start state from the previously observed states, as in Dyna-Q
\cite{sutton1990dynaq}. The advantage of selecting states in this way
is that the distribution or structure of the state space does not need
to be known, but planning may not be guaranteed to cover the state
space entirely. To explore this strategy we experiment with Dyna in
several Atari games.

Dyna is a meta-algorithm. To instantiate it, one must make specific
choices for the model and the value function/policy learner. In the
experiments in this section, the agent uses a perfect model. This
allows us to investigate best case planning performance and focus on
the question of how best to use the model. In Sections
\ref{sec:imperfectmodel} and \ref{sec:online} we incorporate
imperfect, learned models.

We experimented with agents based on two different value function
learners. One was based on the Sarsa algorithm
\cite{rummery1994,sutton1996sarsa}, using linear value function
approximation with Blob-PROST features \cite{liang2016shallow}, which
have been shown to perform well in the ALE. The other was based on
Deep Q-Networks (DQN) \cite{mnih2015human}, which approximates the
value function using a deep convolutional neural
network. Qualitatively, the results were very similar for both agents,
suggesting that our conclusions are robust to different concrete
instantiations of the Dyna architecture. Here we focus on the results
from the DQN-based agent; the results from the Sarsa-based agent can
be found in the appendix.

\subsection{Deep Q-networks and Dyna-DQN}
\label{sec:dynadqn}

Deep Q-networks (DQN) \cite{mnih2015human} is a model-free RL
method, based on Q-learning, that uses a deep convolutional neural
network to approximate the value function.  Unlike Q-learning, DQN
does not update the value function after every step using a single
transition; instead, DQN uses experience replay
\cite{lin1992experience}, and places each observed transition into an
experience replay buffer.  Then, for a single training step, the
algorithm selects a batch of transitions from the buffer uniformly at
random to update the parameters.  Training steps are performed after
every $f$ observed transitions, which is referred to as the
\textit{training frequency}.

It is straightforward to extend DQN to incorporate it into the Dyna
architecture. Our approach, which we call Dyna-DQN, is similar to
other recent attempts to combine DQN with planning
(e.g. \cite{gu2016continuous,kalweit2017uncertainty,peng2018deepdyna}). After
every step taken in the environment, a number of start states for
planning are sampled from a planning buffer containing the agent's
recent real experience.  Keeping a separate buffer ensures that start
states are always from the agent's actual experience.  For each start
state, an action is selected using the agent's current policy and the
model is used to simulate a single transition, which is placed into
the experience replay buffer alongside the transitions observed from
the real environment.  Training continues to happen after every $f$
observed transitions --- real or simulated.  As a result, batches
sampled at training time will contain a mix of real and simulated
experience.

\subsection{Experiments}
\label{sec:dqnvsdyna}
We ran experiments on six games from the ALE and used sticky actions
to inject stocasticity into the emulator (repeat\_action\_probability
= 0.25), as suggested by \citeauthor{machado17arcade}
\shortcite{machado17arcade}.  Each game usually has 18 possible actions,
but some actions are redundant in certain games.  Therefore, we used
the minimal action set like \citeauthor{mnih2015human}
\shortcite{mnih2015human}.

We chose to study the games from the original training set outlined by
\citeauthor{bellemare13arcade} \shortcite{bellemare13arcade}, supplemented
with two additional games, \textsc{Q-Bert} and \textsc{Ms.~Pac-Man},
that \citeauthor{oh2015action} \shortcite{oh2015action} used to evaluate
their model learning approach, which we employ later.  We have omitted
results in \textsc{Freeway} since our implementations of DQN and
Dyna-DQN almost always score zero points at the number of training
frames we used.

Our implementation of DQN used the same hyper-parameters as
\citeauthor{mnih2015human} \shortcite{mnih2015human} with a few small
changes used by \citeauthor{machado17arcade}
\shortcite{machado17arcade}.  At each step, DQN has an
$\epsilon$ probability of selecting a random action instead of the
best action.  We annealed $\epsilon$ from 1.0 to 0.01 over the first
10\% of frames (real and simulated) during learning.  The frame skip
is the number of times a selected action is repeated before a new
action is selected.  We used a frame skip of 5.  Additionally, to
reduce memory use, we used an experience replay buffer size of 500k
transitions instead of the original 1M.

\begin{figure}[tb]
	\centering
	\subfloat{\includegraphics[trim={0 0.1in 0 0.1in},clip,height=0.125in, keepaspectratio]{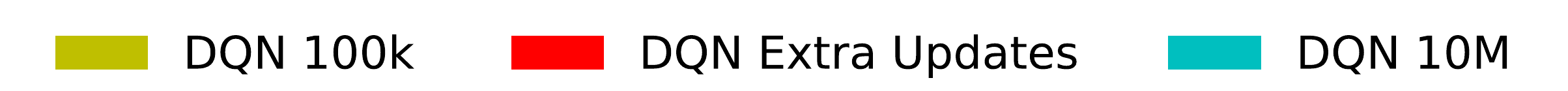}}\\
	\subfloat{\includegraphics[trim={0 0.11in 0 0.11in},clip,height=0.125in, keepaspectratio]{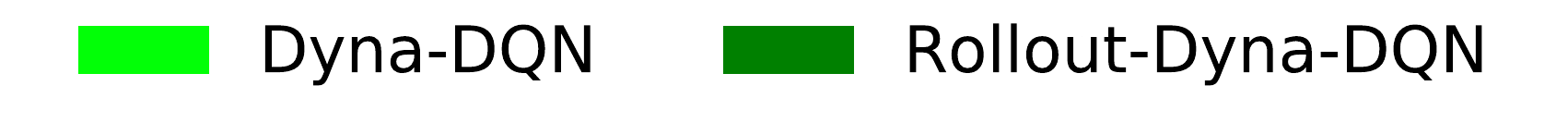}}\\
	\subfloat{\includegraphics[height=1.28in, keepaspectratio]{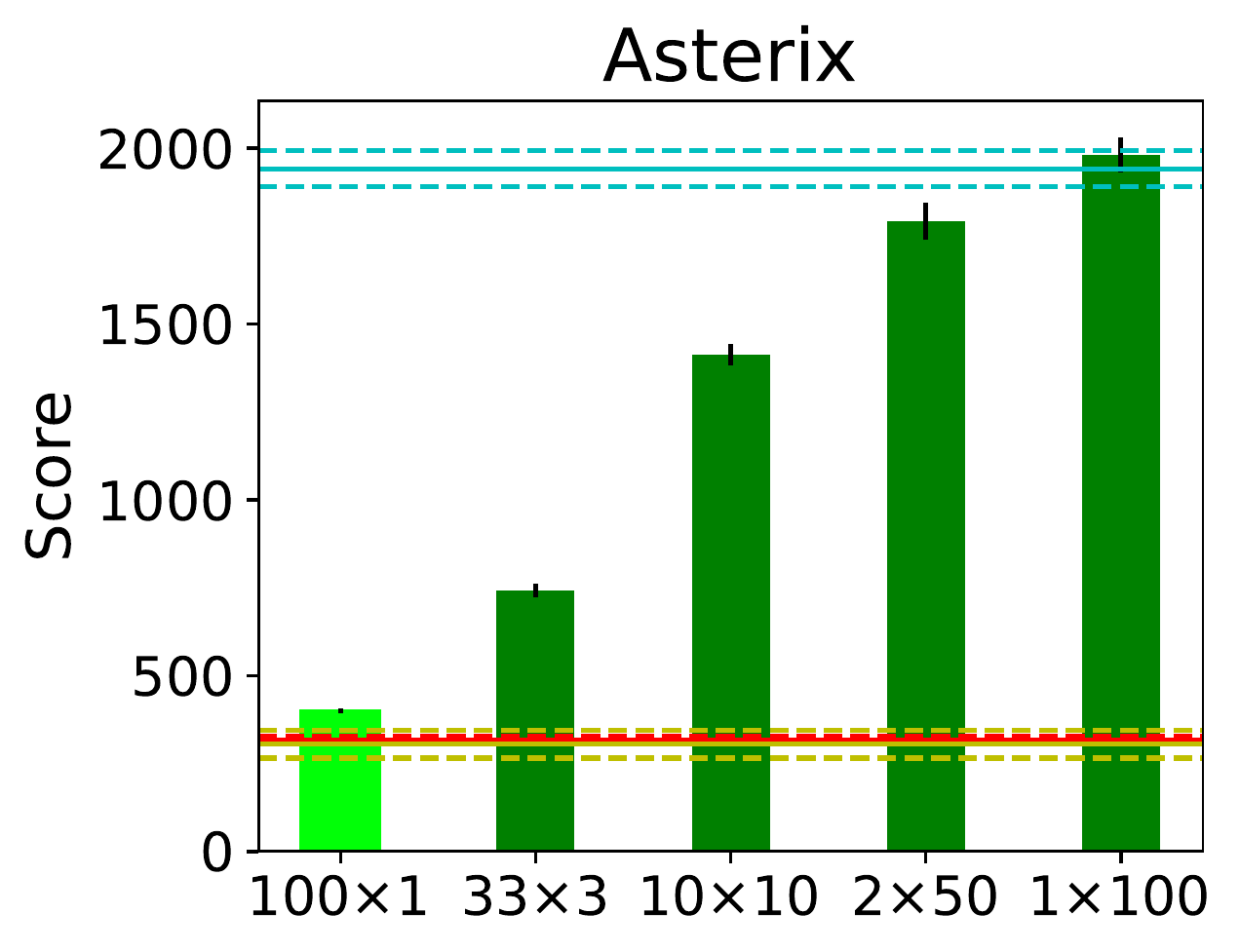}} \hfill
	\subfloat{\includegraphics[height=1.28in, keepaspectratio]{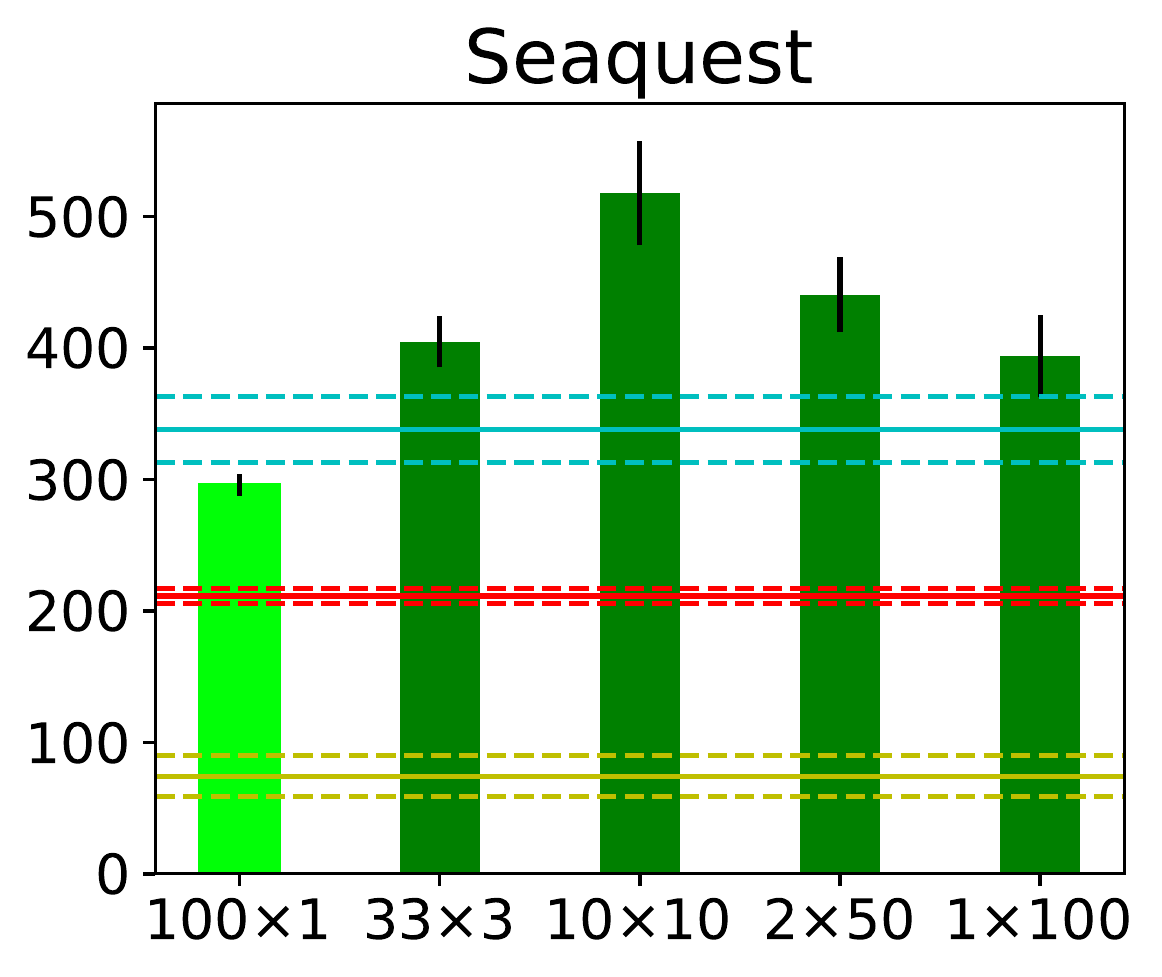}}\\
	\subfloat{\includegraphics[height=1.28in, keepaspectratio]{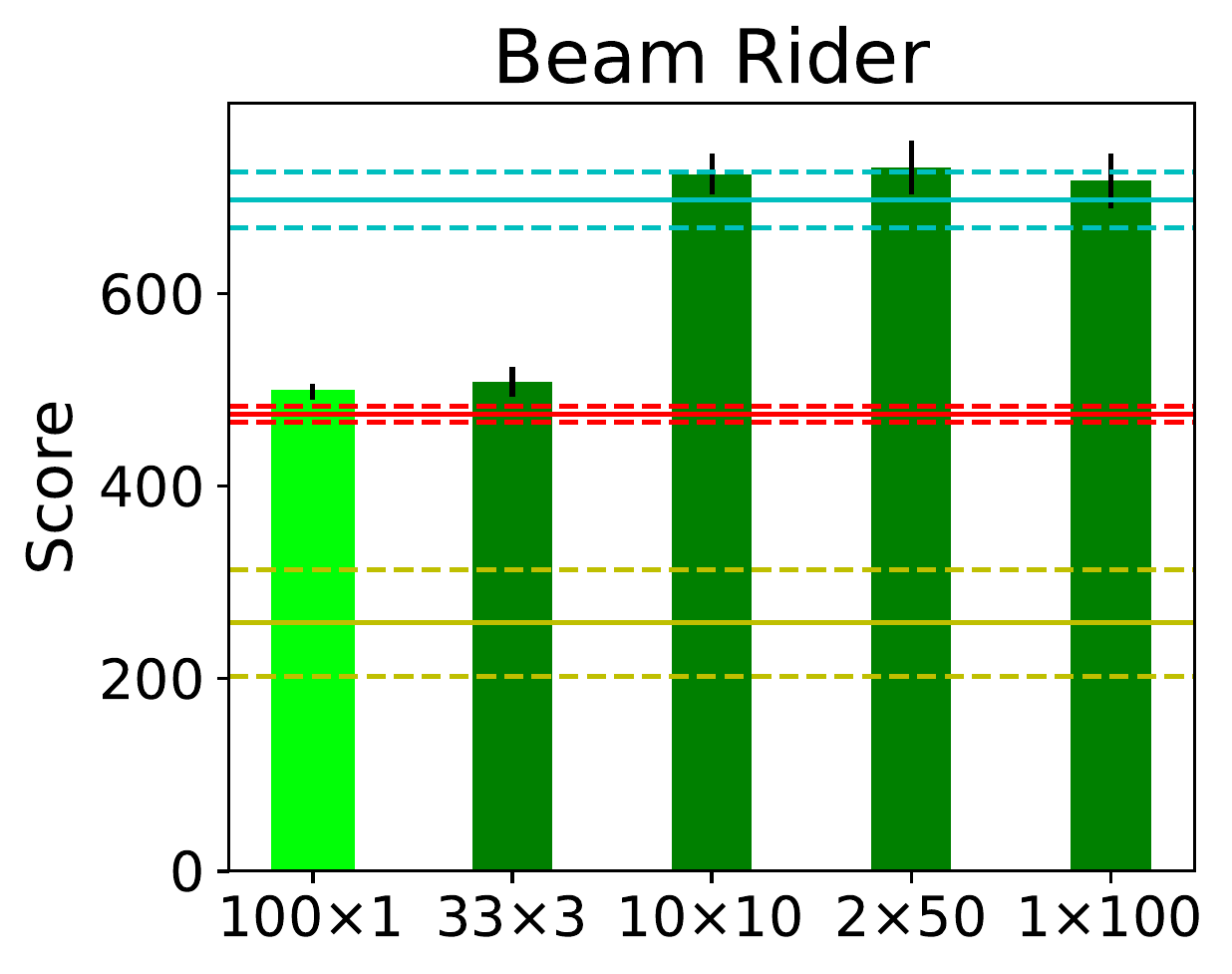}}\hfill
	\subfloat{\includegraphics[height=1.28in, keepaspectratio]{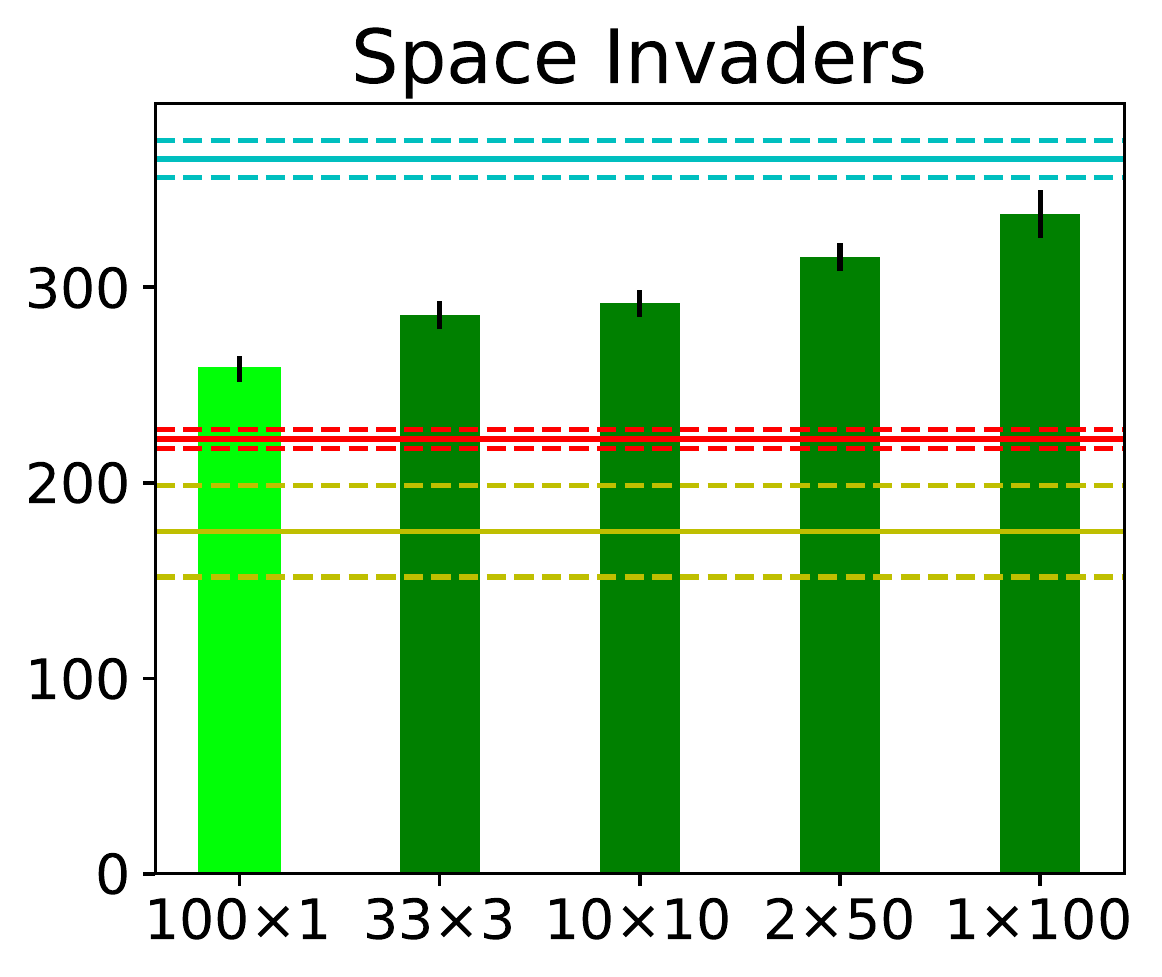}}\\
	\subfloat{\includegraphics[height=1.38in, keepaspectratio]{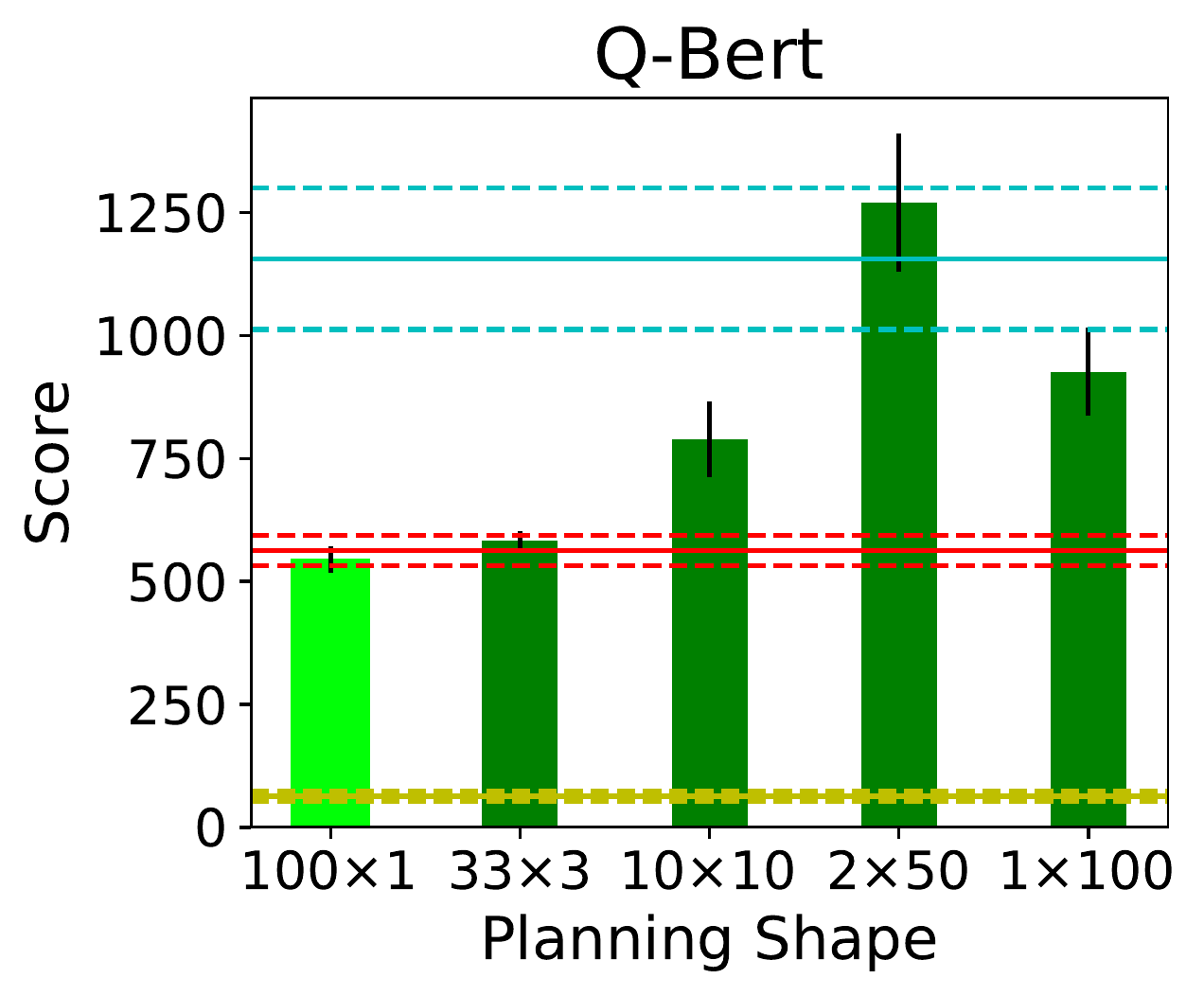}} \hfill
	\subfloat{\includegraphics[height=1.38in, keepaspectratio]{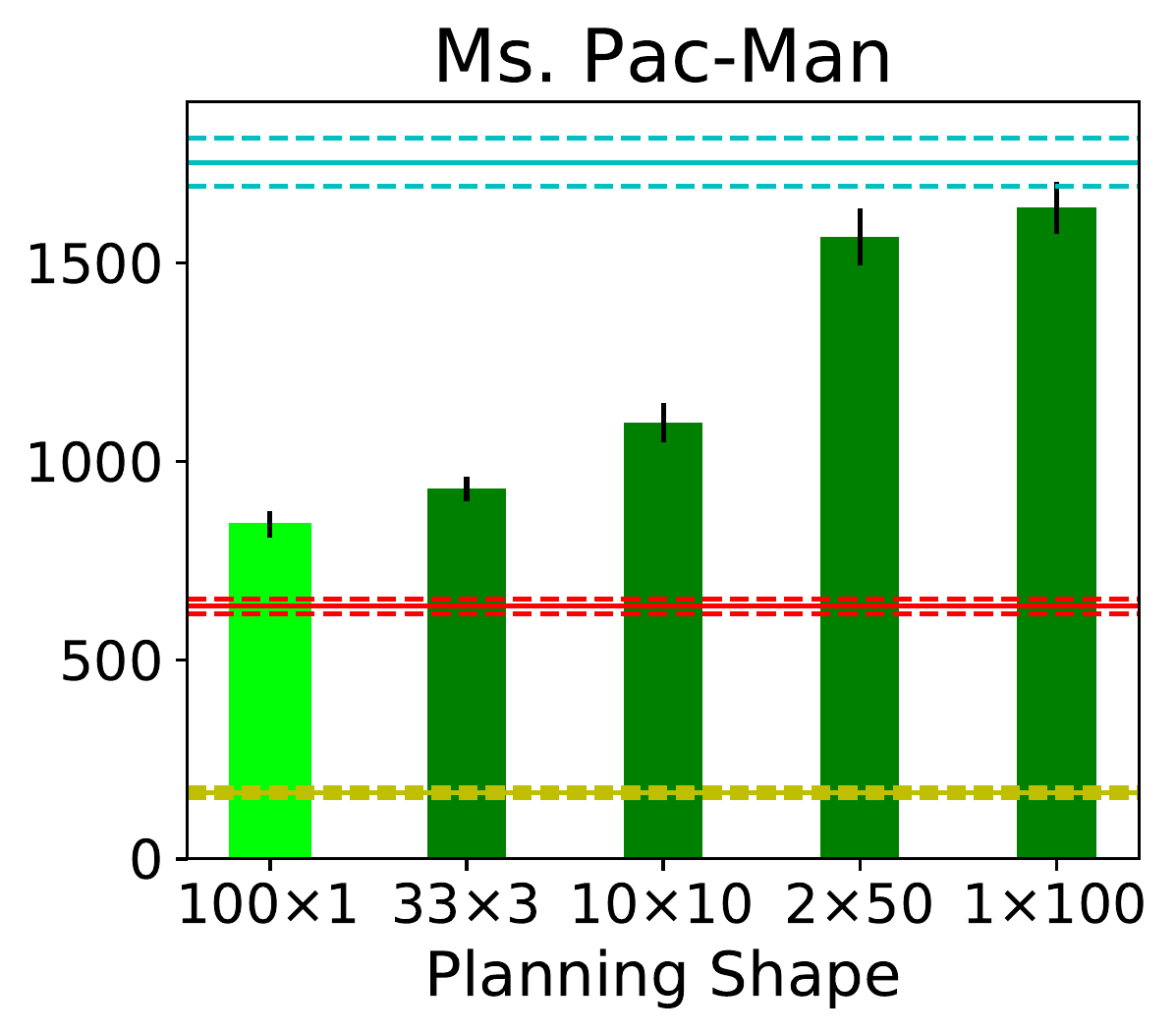}}
	\caption{The results of running Rollout-Dyna-DQN on six games
          from the ALE compared to DQN baselines. The bright green
          bars represent Dyna-DQN (Section \ref{sec:dqnvsdyna}). The
          dark green bars represent Rollout-Dyna-DQN with various
          planning shapes (Section \ref{sec:longerrollouts}). The
          horizontal lines show the baseline scores (see Section \ref{sec:dqnvsdyna}).}
	\label{fig:perfectplots}
\end{figure}

As described above, in these experiments, the agent used a perfect
copy of the emulator for its model.  Start states for planning were
selected from the planning buffer containing the 10,000 most recent
real states observed by the agent, which for all games was multiple
episodes of experience.  For each real step, Dyna-DQN drew 100 start
states from the buffer and simulated a single transition from each.
Dyna-DQN was trained for 100k real frames, or equivalently 10M
combined model and real frames.  The training frequency was every 4
steps of real and model experience.  After training, the mean score in
100 evaluation episodes using a fixed $\epsilon=0.01$ was recorded.
This training and evaluation procedure was repeated for thirty
independent runs.  The mean scores and standard errors for the six
games are shown in Figure~\ref{fig:perfectplots}. The bright green
bars labeled $100 \times 1$ represent Dyna-DQN; the dark green bars
will be described in Section \ref{sec:longer}.

To better evaluate the benefit of model-based updates, we also
compared to the following model-free DQN baselines (pictured as
horizontal lines in Figure~\ref{fig:perfectplots}).

\textbf{DQN 100k:} DQN trained only for 100k real frames (yellow
lines).  This allows us to compare DQN and Dyna-DQN with an equivalent
amount of real experience.  This benchmark serves as a sanity check to
show that using the perfect model to gather additional data does
improve sample complexity.  As expected, Dyna-DQN outperformed DQN
100k; it uses the model to generate more experience and does many more
updates.  However, this benchmark does not indicate whether the
performance increase is due to the additional data generated by
Dyna-DQN, or the extra updates to the value function completed during
planning.

\textbf{DQN Extra Updates:} DQN trained for 100k real frames, but with
the same number of updates to the value function as Dyna-DQN (red
lines).  For each time DQN would normally perform a single training
step, DQN Extra Updates performs 100 training steps.  This way DQN
Extra Updates is like Dyna-DQN, but it uses only experience gathered
from the environment, while Dyna-DQN also generates experience from
the model. DQN Extra Updates allows us to isolate the advantage of
using the model to generate new experience compared to simply doing
more updates with the real experience.  Surprisingly, in every game
excluding \textsc{Seaquest}, Dyna-DQN provided little benefit over DQN
Extra Updates, even with a perfect model.  This indicates that most of
the benefit of planning was from simply updating the value function
more often, which does not require a model.

\textbf{DQN 10M:} DQN trained for 10M frames (cyan lines).  This allows
us to compare DQN and Dyna-DQN with an equivalent amount of total
experience.  We might hope the experience generated by a perfect model
would allow Dyna-DQN to perform comparably to this baseline, but in
most games the performance of Dyna-DQN did not approach that of DQN
10M.  This shows that there is significant room to improve the
performance.  Dyna-DQN and DQN 10M both gather additional data from
the true system and perform the same number of updates; the only
difference is the distribution over the start states of the additional
transitions.

Overall, we find that the extra computation required by Dyna to
utilize the model does not appear to be worth the effort.  Here we
re-emphasize that this finding is robust to different choices for
value function learners within the Dyna architecture; we obtained
similar results using an agent based on Sarsa with linear value
function approximation rather than DQN (see the online appendix). It
seems that planning in this way --- taking a single step from a
previously visited state --- does not provide data that is much
different than what is already contained in the experience replay
buffer.  If true, a strategy is needed to make the data generated by
the model different from what was already experienced.

\section{Planning with Longer Rollouts}
\label{sec:longer}
We hypothesize that it may be possible to generate more diverse
experience by rolling out more than a single step from the start state
during planning. Since the current policy will be used for the
rollout, the model may generate a different trajectory than what was
originally observed.  Longer rollouts would also allow the agent to
see the longer-term consequences of exploratory actions or alternative
stochastic outcomes. In existing work longer rollouts have been
employed in Dyna-style planning
(e.g. \cite{gu2016continuous,kalweit2017uncertainty}), but the impact
of this choice has not been extensively or systematically studied.

It is straightforward to modify Dyna-DQN so that instead of rolling
out a single step, the model is used to roll out $k$ steps from each
start state, producing a sequence of $k$ states and rewards, which are
all placed in the experience replay buffer.  Let this algorithm be
called Rollout-Dyna-DQN.  When $k=1$ we recover exactly Dyna-DQN.

Given a budget of planning time in terms of a fixed number of model
prediction steps, planning could take on a variety of shapes.  Let the
planning shape be described by the notation $n$$\times$$k$,
where $n$ is the number of start states.  For example: 100
rollouts of 1 step (100$\times$1); 10 rollouts of 10 steps
(10$\times$10); or 1 rollout of 100 steps (1$\times$100), each require
the same amount of computation from the model.  The experience
generated during a multi-step rollout is still recorded as single
transitions in the replay buffer, to be sampled independently during a
DQN training step. Furthermore we hold the total number of transitions
drawn from the model constant. Thus the only difference between two
planning shapes is the distribution of transitions in the replay
buffer.  In the next section we investigate the effects of planning
shape on the performance of Dyna-style planning.

\subsection{Experiments}
\label{sec:longerrollouts}
Our experimental setup is the same as in the Section
\ref{sec:dqnvsdyna}, but now the planning shape for Dyna-DQN is
varied.  We trained and evaluated 100$\times$1, 33$\times$3,
10$\times$10, 2$\times$50, and 1$\times$100 planning shapes.  The
results for the six games are shown in dark green in
Figure~\ref{fig:perfectplots}.  The 100$\times$1 planning shape
(bright green) is equivalent to Dyna-DQN.  Note that the ratio of
real to simulated transitions remains the same in each case.  

In every game 100$\times$1 planning achieved the worst performance
whereas longer rollout lengths allowed Rollout-Dyna-DQN to
significantly outperform DQN Extra Updates and approach the
performance of DQN 10M. As before, the positive impact of longer
rollouts is not specific to DQN; similar results were obtained
when repeating the experiment with the Sarsa-based Dyna agent (see the
online appendix). This property of Dyna-style planning seems to be
robust to differing implementation choices.

Again, recall that the only difference between two planning shapes is
the distribution of experience generated by the model.  Thus, our
results suggest that for Dyna to make the most of the model, it is
critical that the model be used to generate sufficiently novel
experience, and generating multi-step rollouts appears to be a simple
and effective strategy for accomplishing this.  Doing longer rollouts
during planning makes using the model worth the effort whereas the
100$\times$1 planning is often no better than doing extra updates with
only real experience.  Also, recall that in these experiments the
agent had access to a perfect model.  With a learned model,
performance will likely be worse due to model errors, so rollouts may
be the only way to obtain a benefit over simply performing extra
updates using the agent's real experience.

\section{Planning with an Imperfect Model}
\label{sec:imperfectmodel}
In the previous sections we have used a perfect model to measure
best-case planning performance. In practice, of course, the agent's
model will typically be learned and thus imperfect. In this section we
replace the perfect copy of the emulator with a learned model
pre-trained on data from expert play.  The goal of this section is not
to evaluate the sample efficiency of Dyna-DQN overall, since
pre-training a model requires data that is not accounted for in the
comparisons.  Instead, the purpose is to investigate the impact of
planning shape on Dyna-style planning when the model is flawed.  In
Section \ref{sec:online} we investigate the case where the model is learned
online, along with the value function.

Previously we have seen that increasing the length of the rollouts
during planning tends to make model-based updates more valuable.
However, since the model is now imperfect, small errors will compound
during long rollouts and make the predictions unreliable
(e.g. \cite{talvitie2014model}).  Therefore, we hypothesize that there
will be competing effects: the algorithm benefits from long rollouts,
but the model's performance degrades as rollout length increases.
This may result in the best performance at some intermediate rollout
length.

\subsection{The Imperfect Model}
For the imperfect model, we made use of the deep convolutional neural
network architecture introduced by \citeauthor{oh2015action}
\shortcite{oh2015action}. This approach has already been shown to learn to
make visually accurate action-conditional predictions for hundreds of
steps on video input from Atari games, so it is a natural choice for
the model-learning component of a Dyna agent in this domain.

The model takes a stack of the four most recent grayscale frames and
the agent's action as input, and outputs a single predicted next
frame. The model can be used to make multi-step predictions by
concatenating the predicted frame with the most recent three history
frames, and running the model forward another step.  To train the
model, \citeauthor{oh2015action} \shortcite{oh2015action} created a
training data set by excuting the policy of a trained DQN agent and
recording the actions and frames.  Then, batches of image histories,
actions, and image targets are drawn and used to train the model to
minimize the average squared error between the predicted and target
frames over $k$-steps.  To increase stability during training a
curriculum approach is used: the model is first trained to make
1-step, 3-step, then 5-step predictions.

In its original formulation the model predicts only the next image,
but an environment model for reinforcement learning needs to predict
both the next state and the next reward.  Therefore, we extend the
model to make reward predictions in a manner similar to
\citeauthor{leibfried2017reward} \shortcite{leibfried2017reward}. In
addition to the input frames, the model is provided the most recent
three reward values as input and predicts the next reward value as
well as the next frame.  Since DQN clips the rewards to the interval
$[-1,1]$, the input and target rewards for the model are also clipped
to the same interval.

Further implementation details may be found in the online
appendix. The goal of these experiments is not to evaluate this
model-learning approach {\em per se}, but rather to use it to supply
an imperfect model with which to study the behavior of Dyna-style
planning. 


\begin{figure}[tb]
	\centering
	\hspace{0.26in}\subfloat{\includegraphics[trim={0 0.25in 0 0.25in},clip, height=0.16in, keepaspectratio]{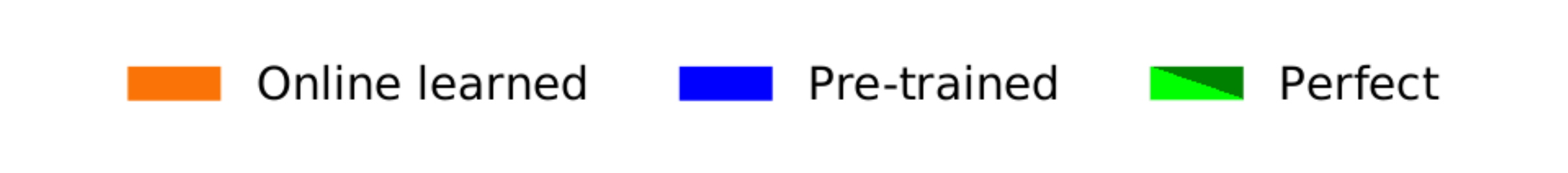}}\\
	\subfloat{\includegraphics[height=1.28in, keepaspectratio]{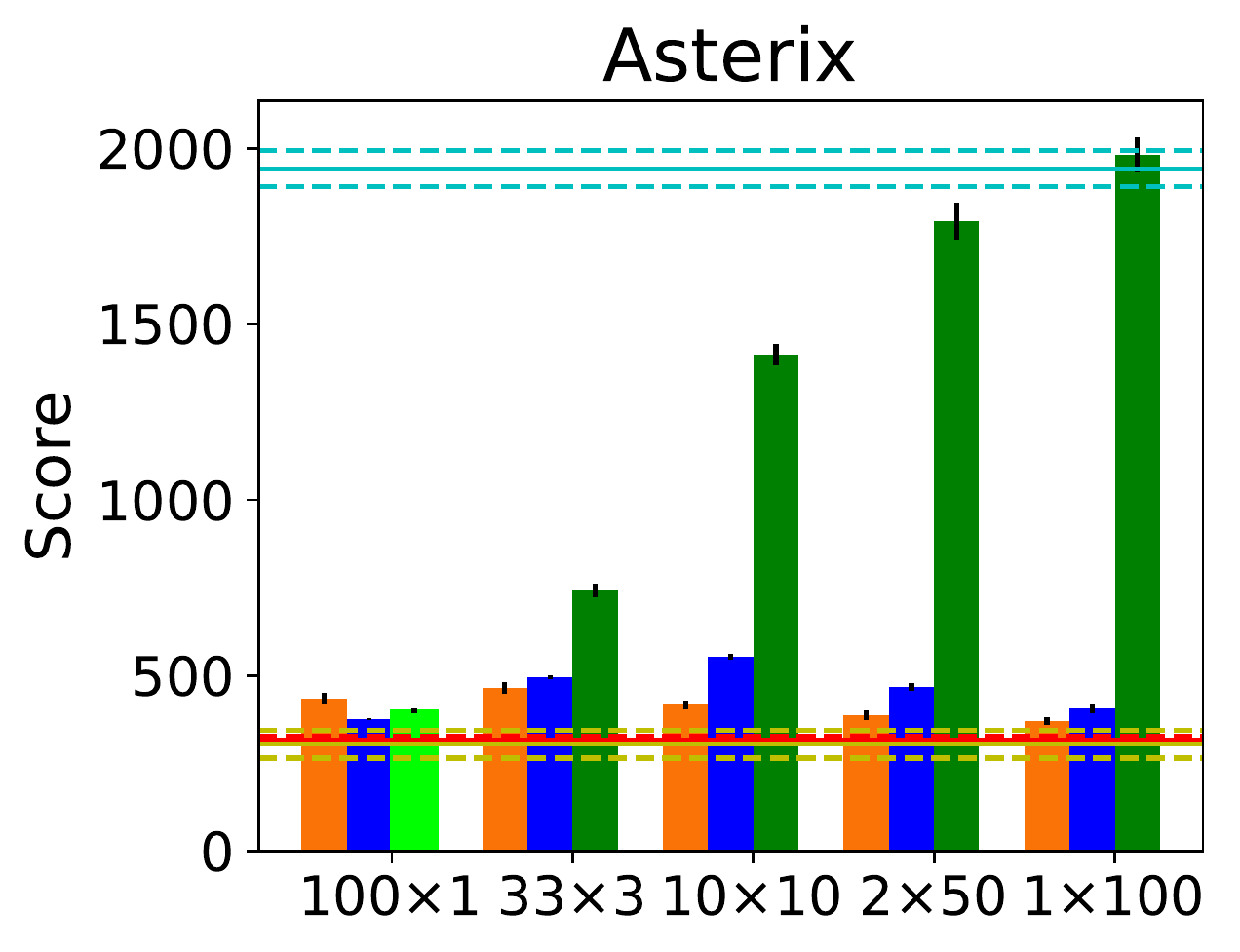}} \hfill
	\subfloat{\includegraphics[height=1.28in, keepaspectratio]{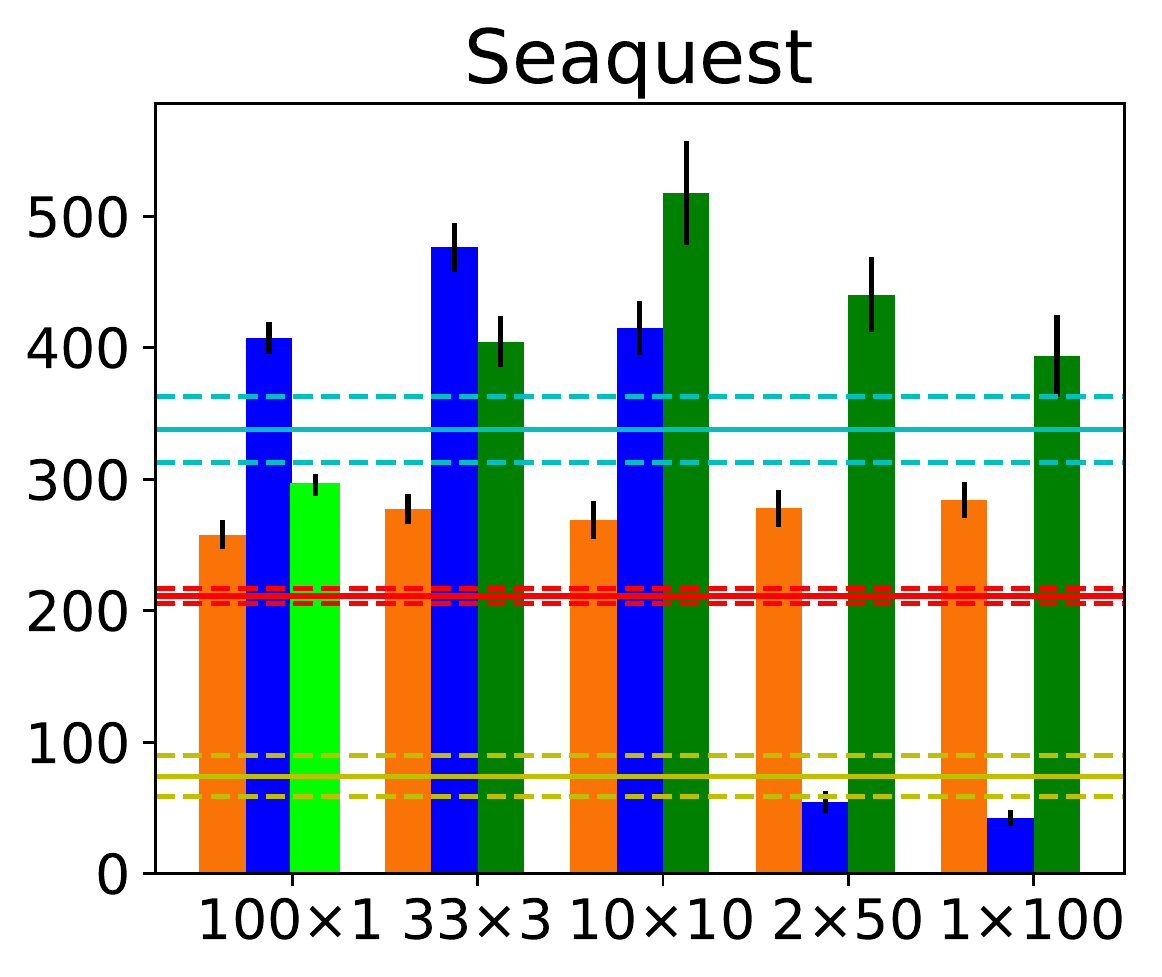}} \\
	\subfloat{\includegraphics[height=1.28in, keepaspectratio]{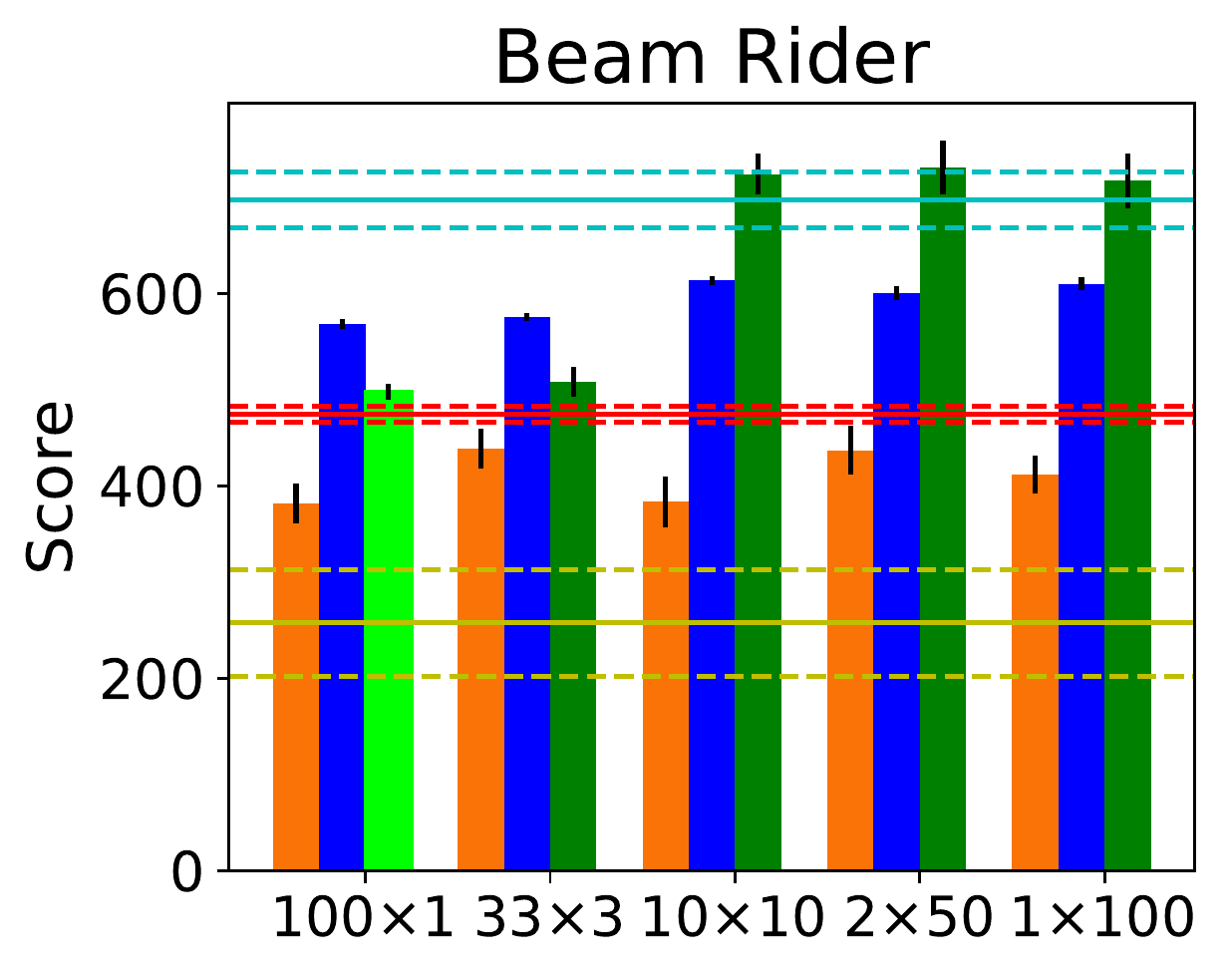}} \hfill
	\subfloat{\includegraphics[height=1.28in, keepaspectratio]{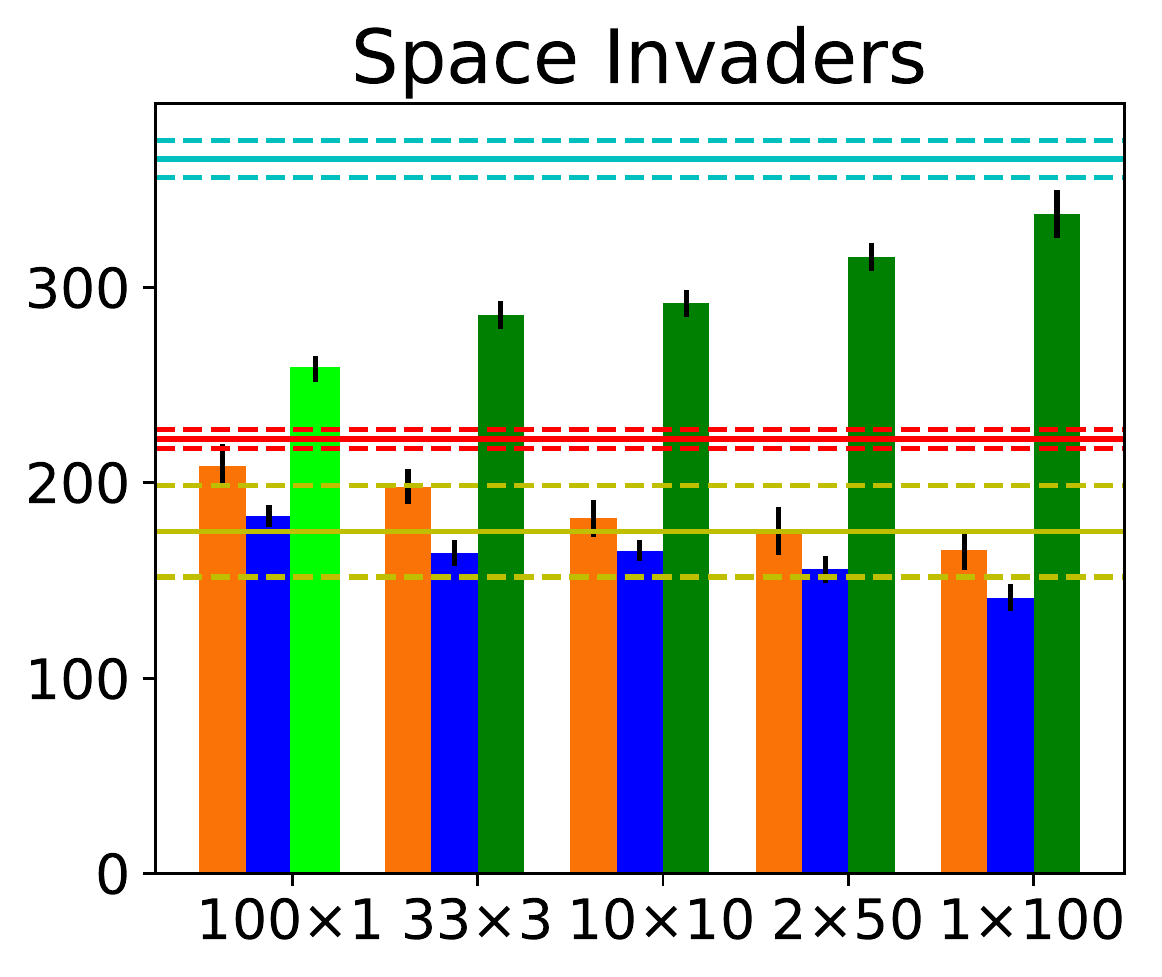}} \\
	\subfloat{\includegraphics[height=1.38in, keepaspectratio]{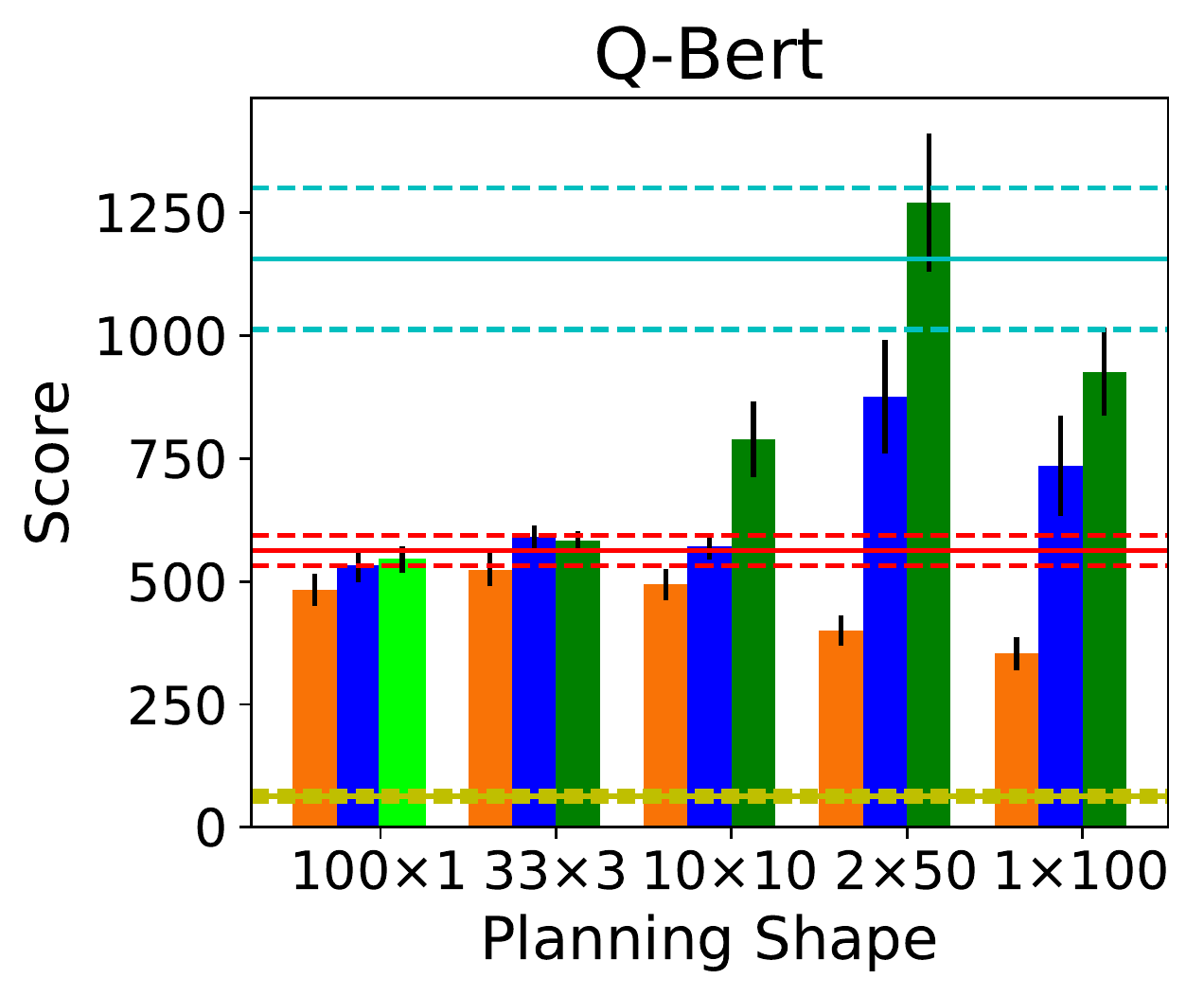}}\hfill
	\subfloat{\includegraphics[height=1.38in, keepaspectratio]{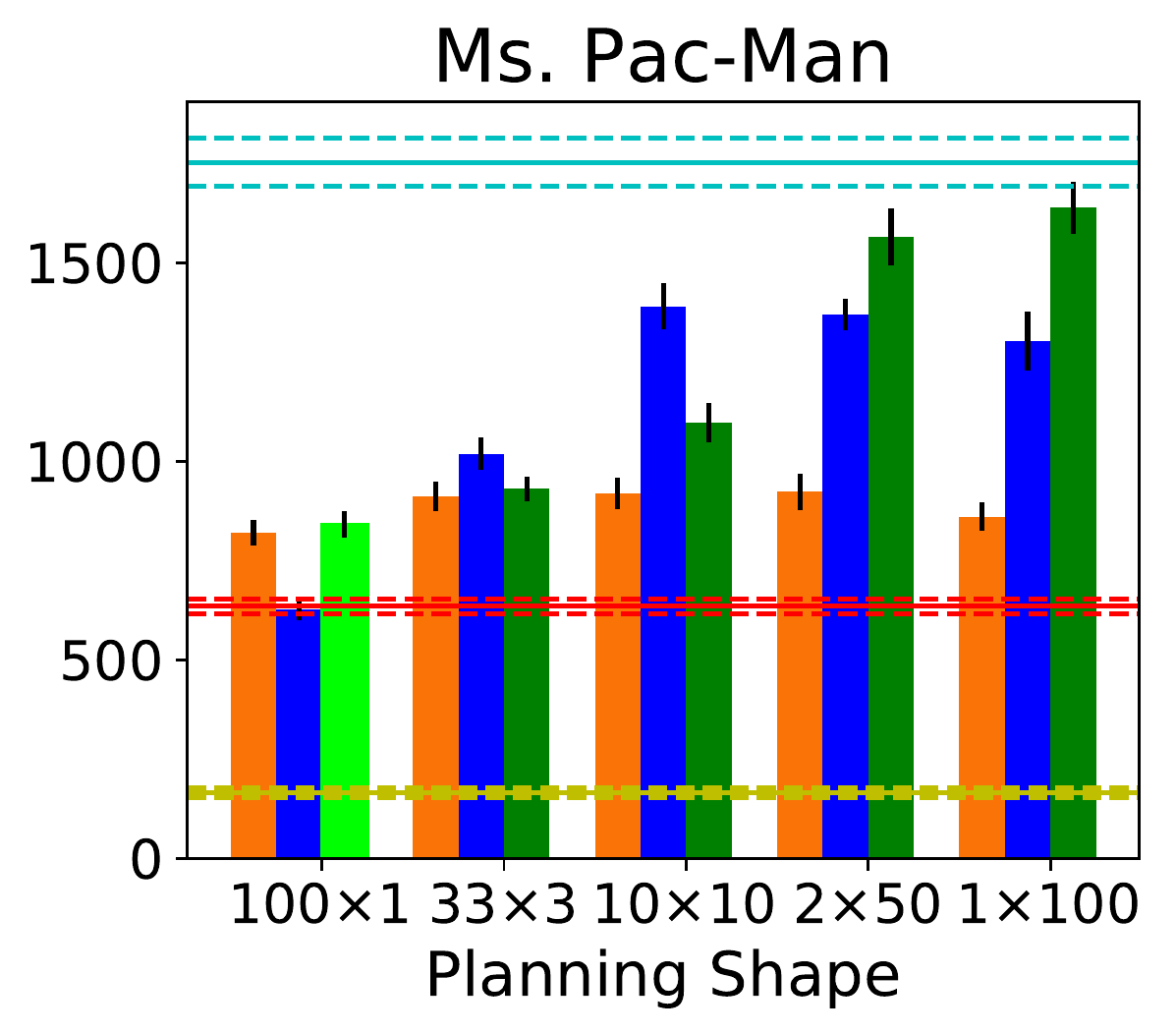}}
	\caption{The results of Rollout-Dyna-DQN with the perfect and imperfect models on six games from the ALE.
		Like the perfect model, using an imperfect model with a rollout length greater than one provides the most benefit.
		The horizontal lines show the same baseline scores as Figure \ref{fig:perfectplots}.}
	\label{fig:learnedplots}
\end{figure}
\subsection{Experiments}
For this section, the experimental setup is the same as in the
previous sections, but the perfect model has been replaced with an
imperfect model. We pre-trained a model with expert data. Then, holding the
model fixed, we repeated the experiment above, measuring
Rollout-Dyna-DQN's performance with various planning shapes. Note that
because the model is pre-trained on a single dataset, our results
cannot be used to draw reliable conclusions about the effectiveness of
this specific model-learning approach.  Our aim is only to study the
impact of model error on Rollout-Dyna-DQN.

The results of applying Rollout-Dyna-DQN with the imperfect,
pre-trained model are shown in Figure~\ref{fig:learnedplots} (blue
bars).  The perfect model results and baselines are the same as in
Figure~\ref{fig:perfectplots}. The orange bars will be described in
Section \ref{sec:onlineexp}.

In every game except \textsc{Space Invaders}, we see evidence that
Rollout-Dyna-DQN with a learned model can outperform DQN Extra Updates
and furthermore that rollouts longer than one step provided the most
benefit.  The reason the performance was especially poor in
\textsc{Space Invaders} is that the model had trouble predicting
bullets, which is fundamental to scoring points.
\citeauthor{oh2015action} \shortcite{oh2015action} attribute this flaw
to the low error signal produced by small objects, which can make it
difficult to learn about small details in the image. In
\textsc{Seaquest} the learned model curiously seems to sometimes
outperform the perfect model. Recall that the learned model is trained
on expert data -- perhaps it is overfitting in a way that happens to
be beneficial to planning.

The results also support our hypothesis that there is a trade-off
between the benefits of long rollouts for planning and the model's
error increasing with rollout length.  For example, in
\textsc{Asterix} the performance peaked at 10x10 planning and dropped
off as rollouts became shorter or longer.  Similarly, in most other
games the best-performance was achieved at medium rollout length. In
order to ensure that these observations are not specific to this
particular learned model, we trained two additional models using the
same neural net architecture but different training
protocols. Qualitatively we observed similar phenomena: planning shape
plays a significant role in performance and trades off with model
error in long rollouts. The details and results from those experiments
may be found in the online appendix.

The trends are certainly not as clear as with the perfect model;
because of the tradeoff with model accuracy, the optimal rollout
length will certainly depend on both the problem and the modeling
approach. Nevertheless the conclusion remains: planning shape can
significantly impact performance and is thus an important
consideration for Dyna agents.

\section{Planning and Learning Online}
\label{sec:online}
In all the experiments so far, a perfect model or a pre-trained
learned model has been used; the obvious next step is to study
Rollout-Dyna-DQN in the case where the model is learned alongside the
value function.  Learning the model and value function together
adheres to the original conception of Dyna, and also allows us to
investigate whether learning a model can improve sample efficiency in
this context.

It should be noted that learning the model online raises a host of
complex concerns
(\citeauthor{ross2012agnostic}~\shortcite{ross2012agnostic} offer a
theoretical discussion of some of these issues). For instance,
exploration becomes especially critical, lest initial model errors
cause agent behavior that fails to visit states where the errors occur
(ensuring that they will never be corrected). Also, since both the
model and the behavior policy are constantly changing, they effect one
another in complex and non-stationary ways (this may at times be
advantageous or disadvantageous). In this work we do not attempt to
substantively address these issues. We primarily seek to evaluate the
impact of planning shape on a full-fledged Dyna agent. In so doing, we
may also gain insight regarding the viability of this approach with a
relatively simple agent architecture.

\subsection{Experiments}
\label{sec:onlineexp}
The experimental setup was the same as in the previous experiments
with Rollout-Dyna-DQN, except the model was learned online alongside
the value function (training details are available in the online
appendix). The results for the six games compared to the pre-trained
and perfect models are shown in Figure~\ref{fig:learnedplots} (orange
bars). The main observation is that, as with the perfect and
pre-trained models, the best performance was achieved using rollouts
longer than one step (except in \textsc{Space Invaders}).

Furthermore, though performance with the online-learned model is
generally worse than with the pre-trained model, in three games
(\textsc{Asterix}, \textsc{Seaquest}, and \textsc{Ms. Pac-Man}), it
was consistently above DQN Extra Updates. Thus, despite the potential
pitfalls, we observe that in some cases there is a modest sample
efficiency advantage to learning and planning with a dynamics model
online over re-using the agent's prior experience. To our knowledge,
this is the first time that this has been demonstrated in the ALE,
which has proven to be very challenging for model-based approaches
\cite{machado17arcade}. While this specific instantiation of Dyna is
not a plausible competitor to successful model-free agents in this
domain, our results suggest that Dyna-style planning is a viable
approach that, with further study, may lead to more successful
model-based agents. Importantly, the key finding of this paper is
that, even if model quality is significantly improved, planning shape
must be a consideration in order to get the most out of the learned
model.

\section{Conclusions and Future Work}

Despite the introduction of increasingly effective approaches for
learning predictive models in Atari Games
\cite{bellemare2013bayesian,bellemare2014skip,oh2015action}, this is
the first time that a sample efficiency benefit has been obtained from
learning and planning with a dynamics model in this challenging
domain.  The results show that Dyna is a promising approach for
model-based RL in high-dimensional state spaces. However, we also
found that the benefit of Dyna-style planning is limited when
model-generated experience takes only one step away from the agent's
stored experience, as in the original Dyna-Q. To get the most value
from model-based updates, the model must be used to generate novel
experience. We found that using the model to generate fewer, longer
rollouts was an effective way to achieve this. This finding seems to
be robust across different concrete instantiations of Dyna --- our
results indicate that planning shape has a notable impact on the
benefit of planning with different value learners, multiple
pre-trained models, and a model learned online along with the value
function. The findings in this work also suggest multiple next steps.

We found that the optimal rollout length was unpredictable with
imperfect models, as it depends on the model's reliability in long
rollouts. Our results suggest that it would be valuable to develop
methods that could adaptively select planning shape, perhaps by
monitoring model accuracy in some way.

Some of the model flaws observed by \citeauthor{oh2015action}
\shortcite{oh2015action} were harmful for planning; perhaps
improvements in architecture could benefit MBRL performance.  Our
results show, however, that the benefits of more accurate models will
be most apparent when using longer rollouts --- even with a perfect
model the value of one-step planning is limited.  

In the experiments, start states for planning were selected from a
buffer containing the agent's recent real history; it would be
interesting to generate promising or interesting start states that may
not have been visited by the agent.  This would likely involve
learning a generative model of the states, which might be accomplished
with a solution like a variational autoencoder \cite{kingma2013vae} or
a generative adversarial network \cite{goodfellow2014gan}.

Finally, though longer rollouts were found to be an effective way to
use the model to generate experience, there are other promising
approaches. For instance one might use inverse dynamics models
\cite{pan2018er,goyal2018recall} to effectively propagate value
updates backwards in a manner similar to prioritized sweeping
\cite{moore1993prioritized,peng1993prioritized}.  It may be possible
to combine these insights, exploiting a forward model's ability to
reveal novel states and a backward model's ability to efficiently
improve the value function.

\bibliographystyle{named}
\bibliography{rolloutDyna_arxiv.bib}

\appendix
\renewcommand\thefigure{\thesection.\arabic{figure}}
\section{Rollout-Dyna-Sarsa Results}
In addition to the experiments with DQN and Rollout-Dyna-DQN, we also conducted similar experiments using Sarsa \cite{rummery1994,sutton1996sarsa} with linear function approximation and Blob-PROST features \cite{liang2016shallow}.
We used the same hyper-parameters as \citeauthor{liang2016shallow} \shortcite{liang2016shallow}, except we set $\lambda=0$, instead of $\lambda=0.9$, to better isolate the effects of planning shape from any interactions with eligibility traces.

\begin{figure}[h!]
	\centering
	\subfloat{\includegraphics[height=1.28in, keepaspectratio]{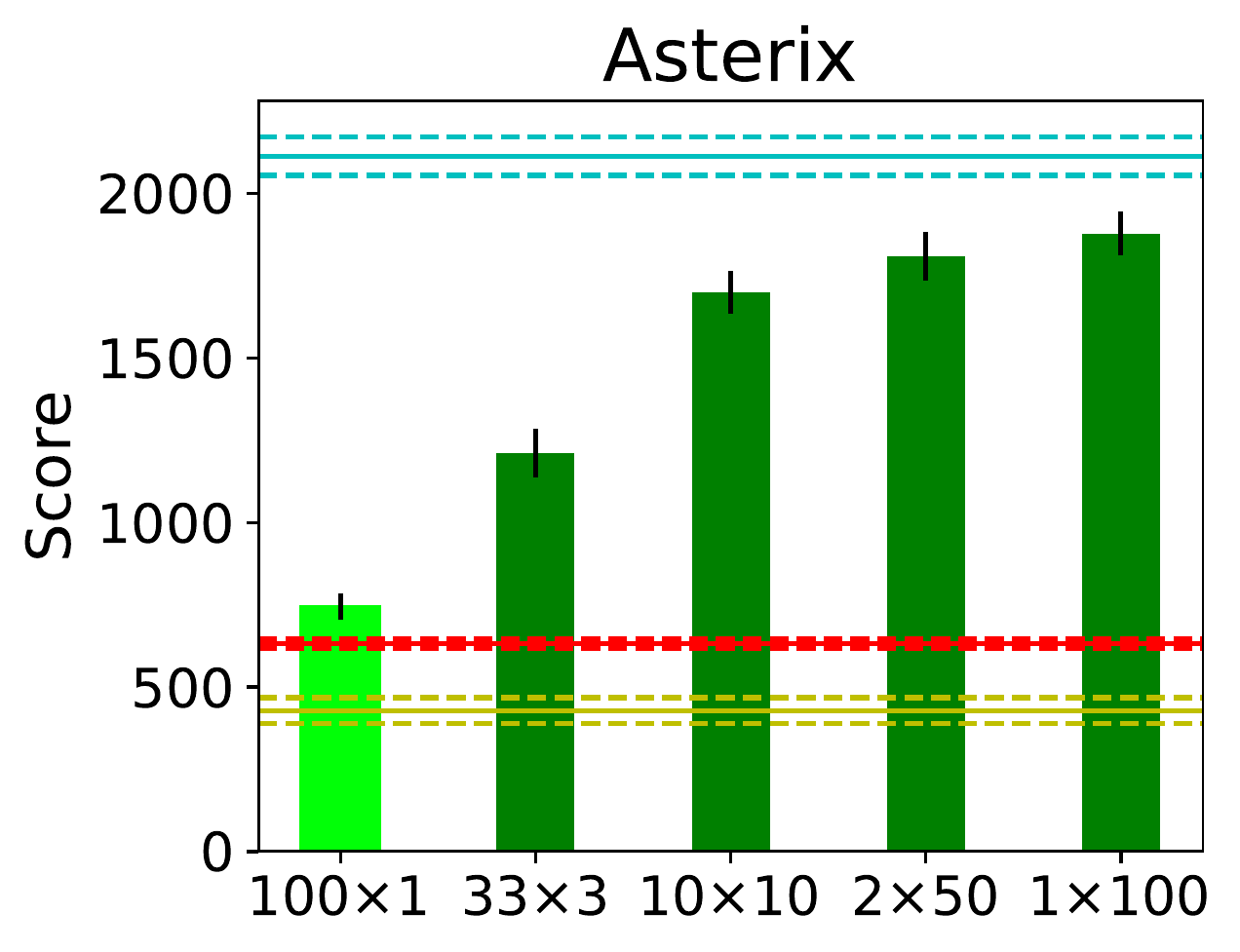}}
	\hfill 
	\subfloat{\includegraphics[height=1.28in, keepaspectratio]{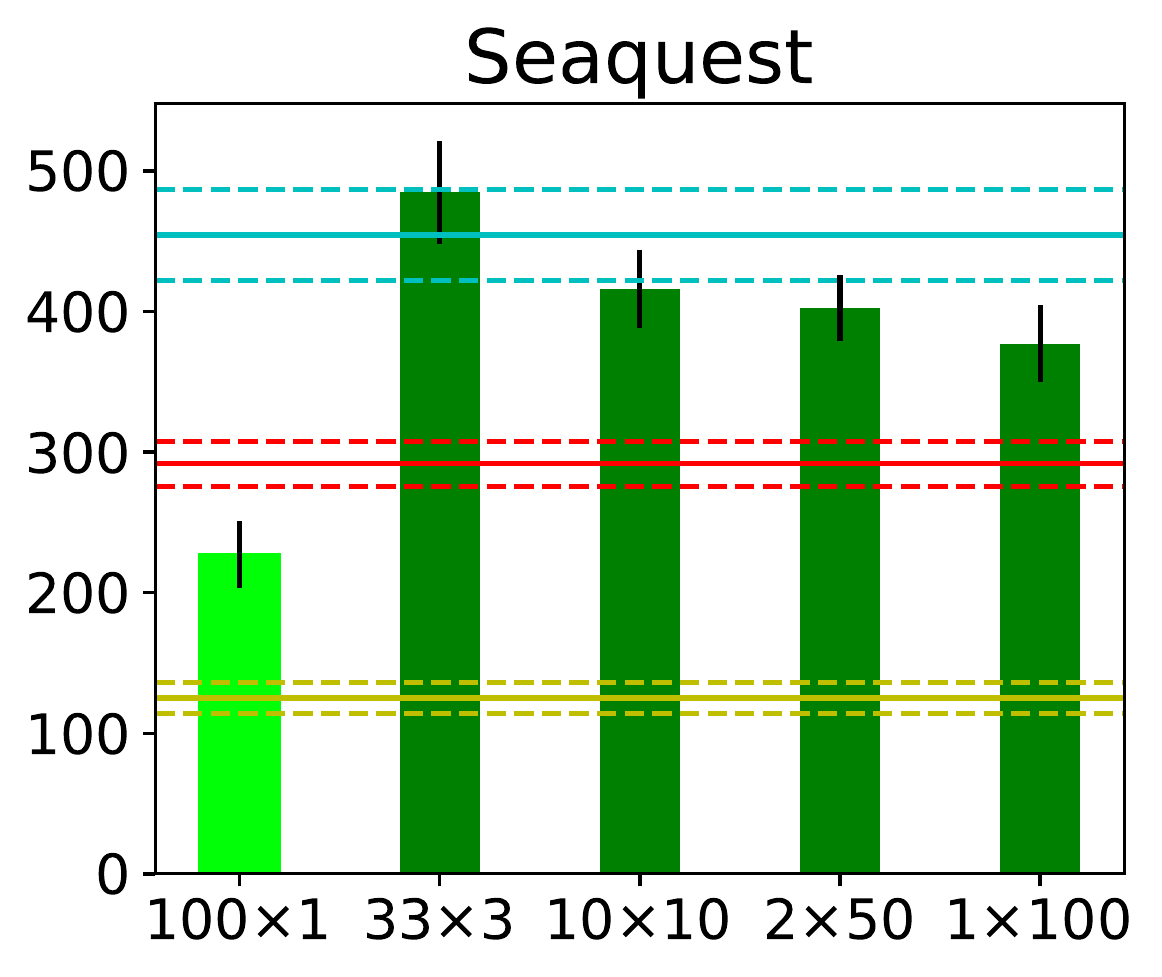}}\\
	\subfloat{\includegraphics[height=1.28in, keepaspectratio]{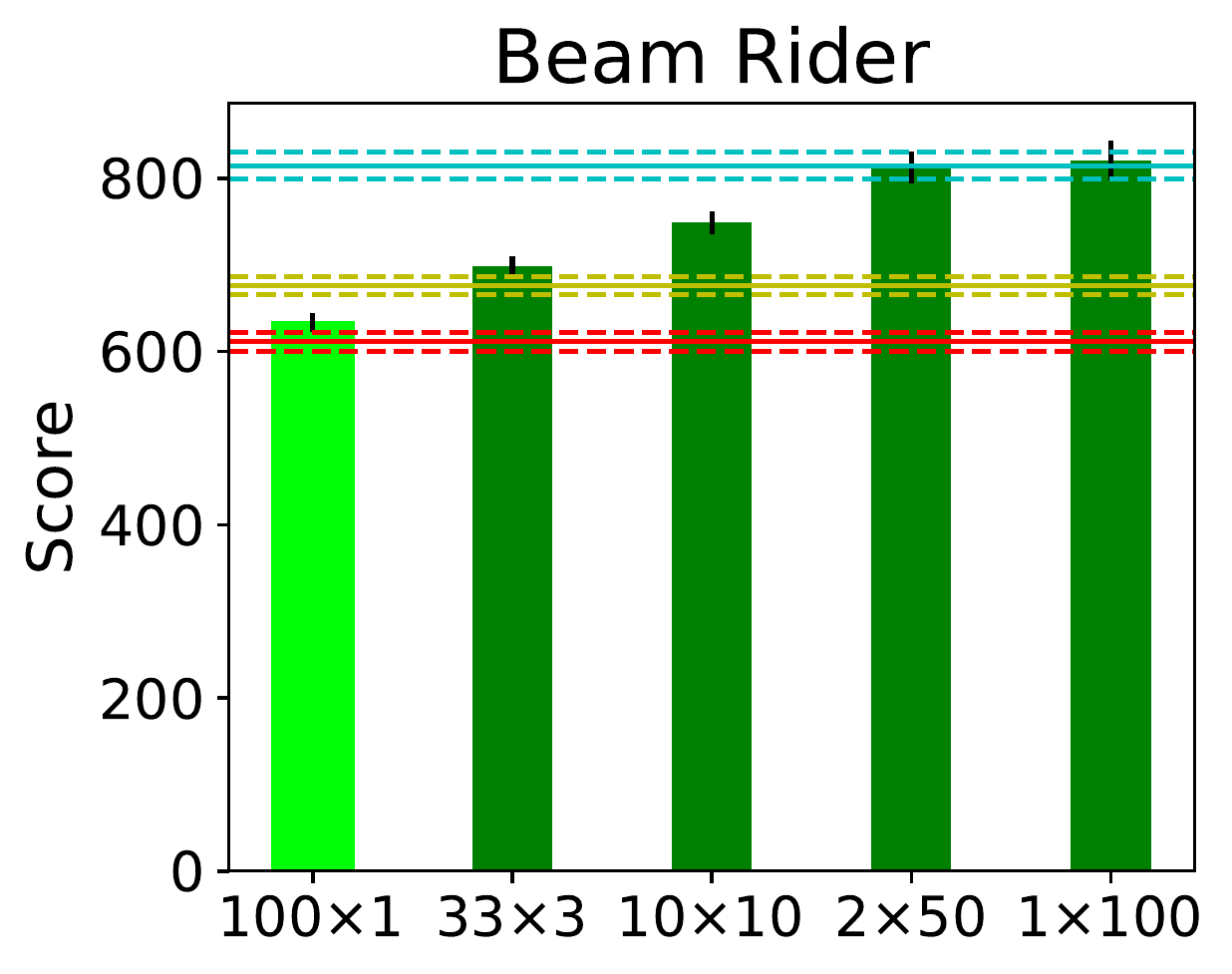}}
	\hfill
	\subfloat{\includegraphics[height=1.28in, keepaspectratio]{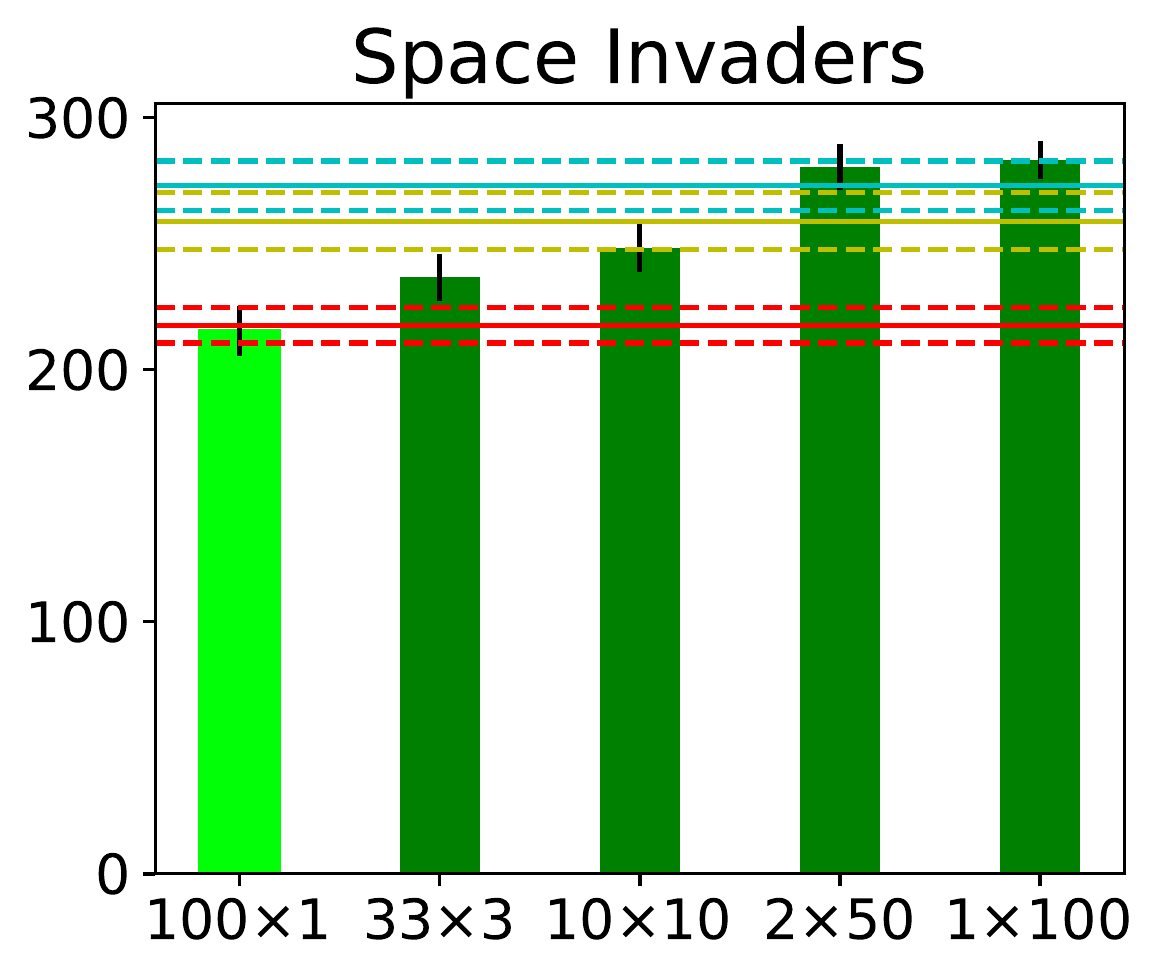}}\\
	\subfloat{\includegraphics[height=1.38in, keepaspectratio]{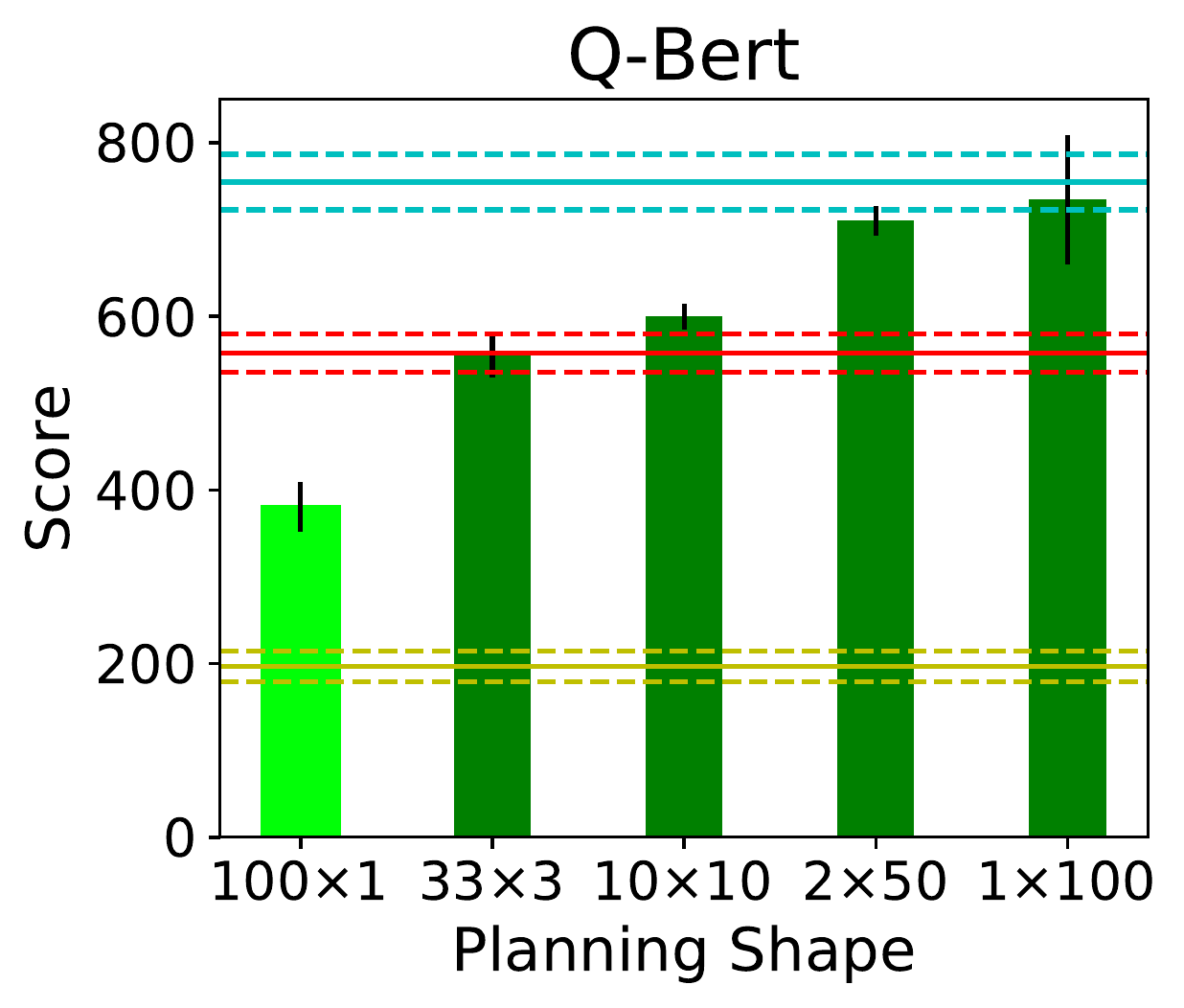}}
	\hfill
	\subfloat{\includegraphics[height=1.38in, keepaspectratio]{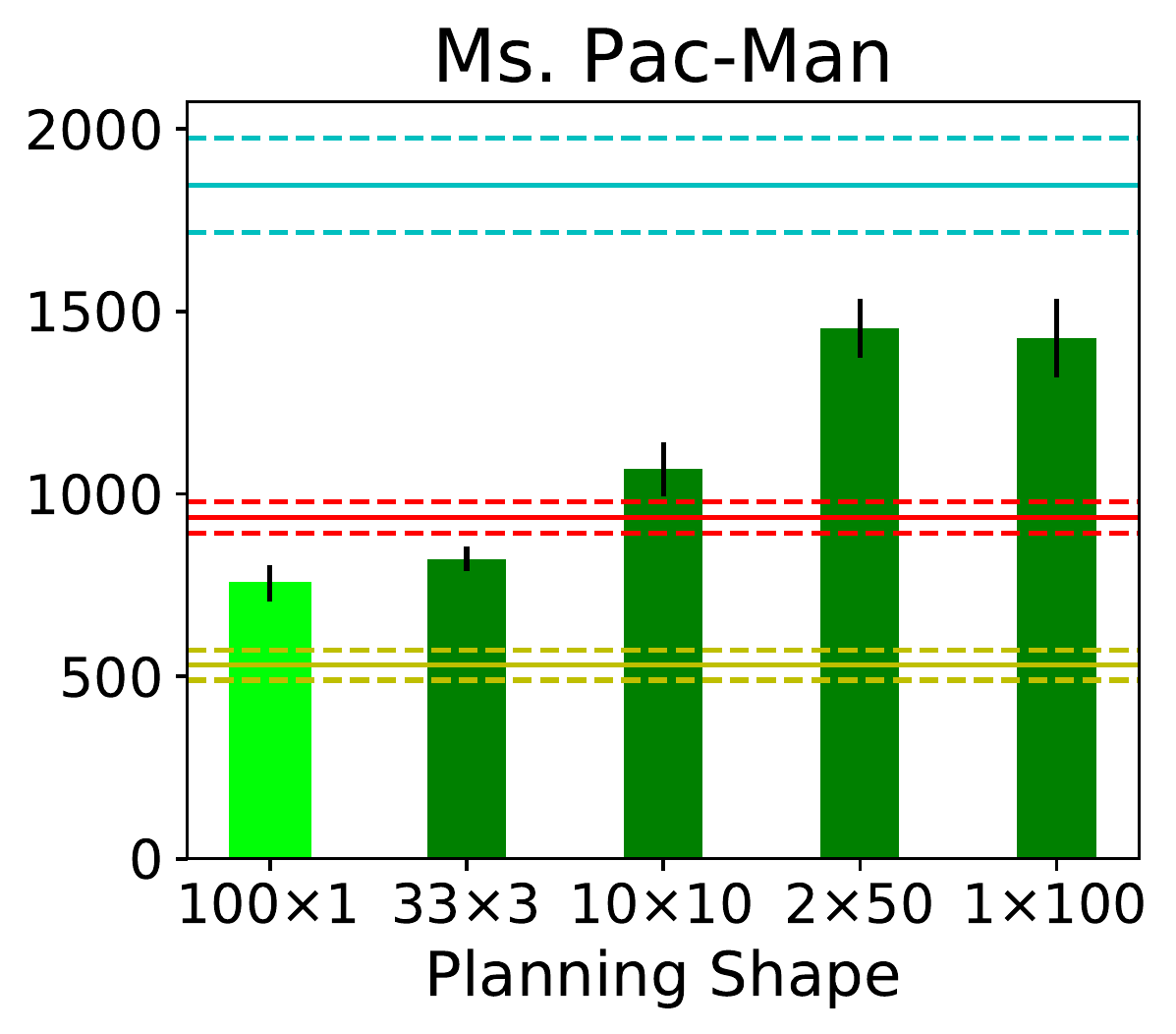}}
	\caption{The results of running Rollout-Dyna-Sarsa on six games from the ALE compared to the baselines.
		There is a planning shape with a rollout length greater than one that outperforms both 100$\times$1 and Sarsa Extra Updates (red line) across all the games.
	}
	\label{fig:perfectsarsaplots}
\end{figure}

We implemented Rollout-Dyna-Sarsa by performing $n$ rollouts, of $k$ steps, after every real step taken by the agent.
After every step of the rollout, the value function was updated using the normal Sarsa update rule.
The start states for planning were selected uniformly randomly from the 10k most recent states experienced by the agent.
For these experiments we assumed the agent had access to a perfect model.
The results are shown in Figure~\ref{fig:perfectsarsaplots}. 
The reported scores are the mean for each algorithm in 100 evaluation episodes after learning for 10M combined real and model frames, and are an average of thirty independent runs.

The model-free baselines that we compared to are similar to the ones used for Rollout-Dyna-DQN.
Sarsa 100k (yellow line in Figure~\ref{fig:perfectsarsaplots}) is a Sarsa agent trained for 100k real frames. 
Sarsa Extra Updates (red line in Figure~\ref{fig:perfectsarsaplots}) is the same as Sarsa 100k, except after every real step it does 100 extra updates using experiences sampled from the agent's recent history.
Sarsa 10M (cyan line in Figure~\ref{fig:perfectsarsaplots}) is a Sarsa agent trained for 10M real frames.

Similar to Rollout-Dyna-DQN, 100$\times$1 planning failed to outperform Sarsa Extra Updates. 
Also, in every game there was a planning shape with a rollout length greater than one that outperformed both 100$\times$1 planning and Sarsa Extra Updates.
These results demonstrate that this phenomenon is not specific to only DQN.

\section{Imperfect Model Details}
\begin{figure*}[t]
	\centering
	\includegraphics[width=0.8\linewidth]{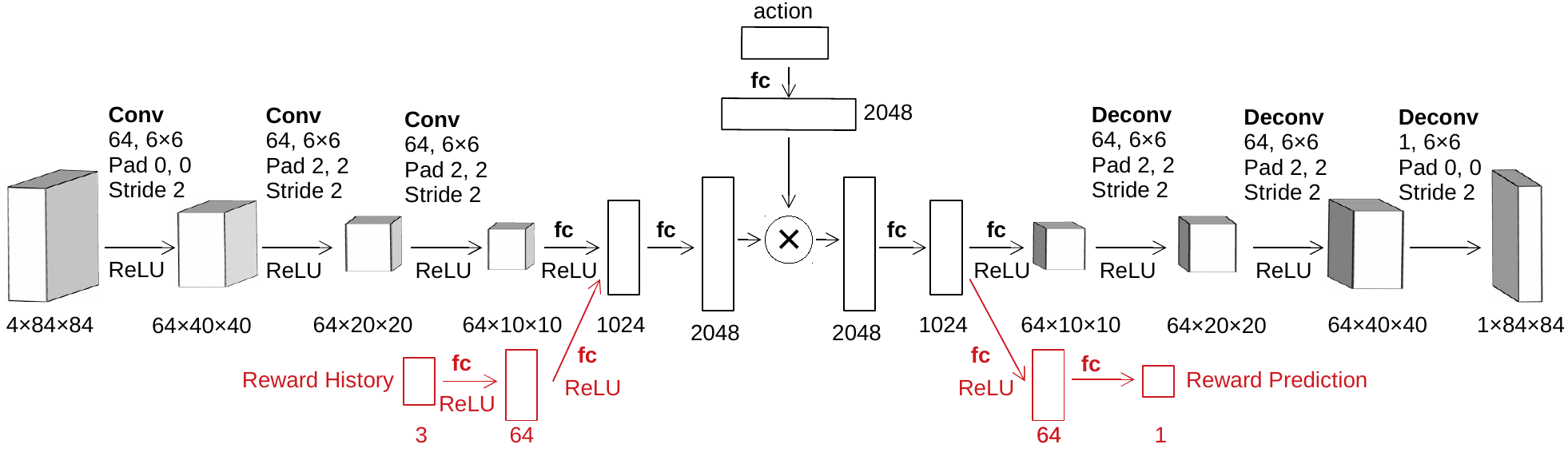}
	\caption{\protect\citeauthor{oh2015action}'s \protect\shortcite{oh2015action} action-conditional video prediction model (black) and its extension to predict the next reward (red).}
	\label{fig:actionconditional}
\end{figure*}
This appendix provides additional details about the deep-neural network that was used for the imperfect model, used in Sections 4 and 5.
The architecture is based on the one introduced by \citeauthor{oh2015action} \shortcite{oh2015action}.
This model was shown to make visually accurate predictions for hundreds of steps on video input from Atari games conditioned on actions.

The model encodes a stack of four grayscale frames into a feature vector using a series of convolutional layers.
The effect of the action 
is applied via a multiplicative interaction between the feature vector and the action.
After the action transformation, the resulting vector is decoded using a series of deconvolutions before finally outputting the single next frame.
The model can be used to make $k$-step predictions by concatenating the predicted frame with the most recent three history frames, and running the model forward another step.
A diagram of the model is shown in Figure \ref{fig:actionconditional} (in black).

To train the model, \citeauthor{oh2015action} \shortcite{oh2015action} created a training data set by running a trained DQN agent and recording the actions and frames.
Then, batches of image histories, actions, and image targets are drawn and used to train the model to minimize the average squared error between the predicted and target frames (denoted $\hat{\mathbf x}$ and $\mathbf x$ respectively) over the $k$-steps. 
To increase stability during training a curriculum approach is used:
the model is first trained to make 1-step, 3-step, then 5-step predictions.

In its original formulation the model predicts only the next state, but an environment model for reinforcement learning needs to predict both the next state and the next reward.
Therefore, we extend the model to make reward predictions by adding a separate fully connected layer after the action transformation, followed by an output layer that predicts a single scalar reward (shown in red in Figure~\ref{fig:actionconditional}).
Thus, in addition to the $k$-step image reconstruction loss, the model is trained to minimize the $k$-step squared difference between the predicted, and target rewards ($\hat r$ and $r$ respectively):
\begin{equation}
\mathcal L_k(\boldsymbol\theta)=\frac{1}{2k} \sum^k_{\kappa=1} \left(\norm{\mathbf{\hat x}_\kappa-\mathbf{x}_\kappa}^2 + \norm{\hat r_\kappa - r_\kappa}^2 \right). 
\end{equation}
This approach is similar to what was used by \citeauthor{leibfried2017reward} \shortcite{leibfried2017reward} to jointly predict frames and rewards.
As input, we also provide the reward history for the three transitions associated with the input frames.
After the reward history input layer, there is a fully connected layer, before joining with the output of the encoder at the action transformation.
Since DQN clips the rewards to the interval $[-1,1]$, the input and target rewards for the model are also clipped to the same interval.
We used this new architecture to train three different models for each game.

\section{Details of Learned Model Training and Additional Results}
\begin{figure}[h!]
	\centering
	\hspace{0.26in}\subfloat{\includegraphics[trim={0 0.25in 0 0.25in},clip, height=0.16in, keepaspectratio]{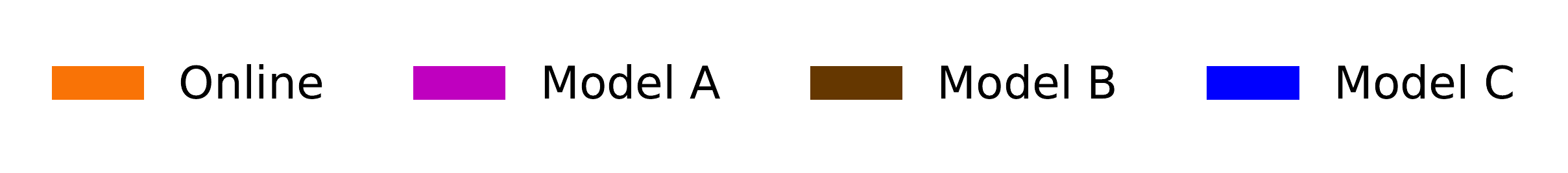}}\\
	\subfloat{\includegraphics[height=1.28in, keepaspectratio]{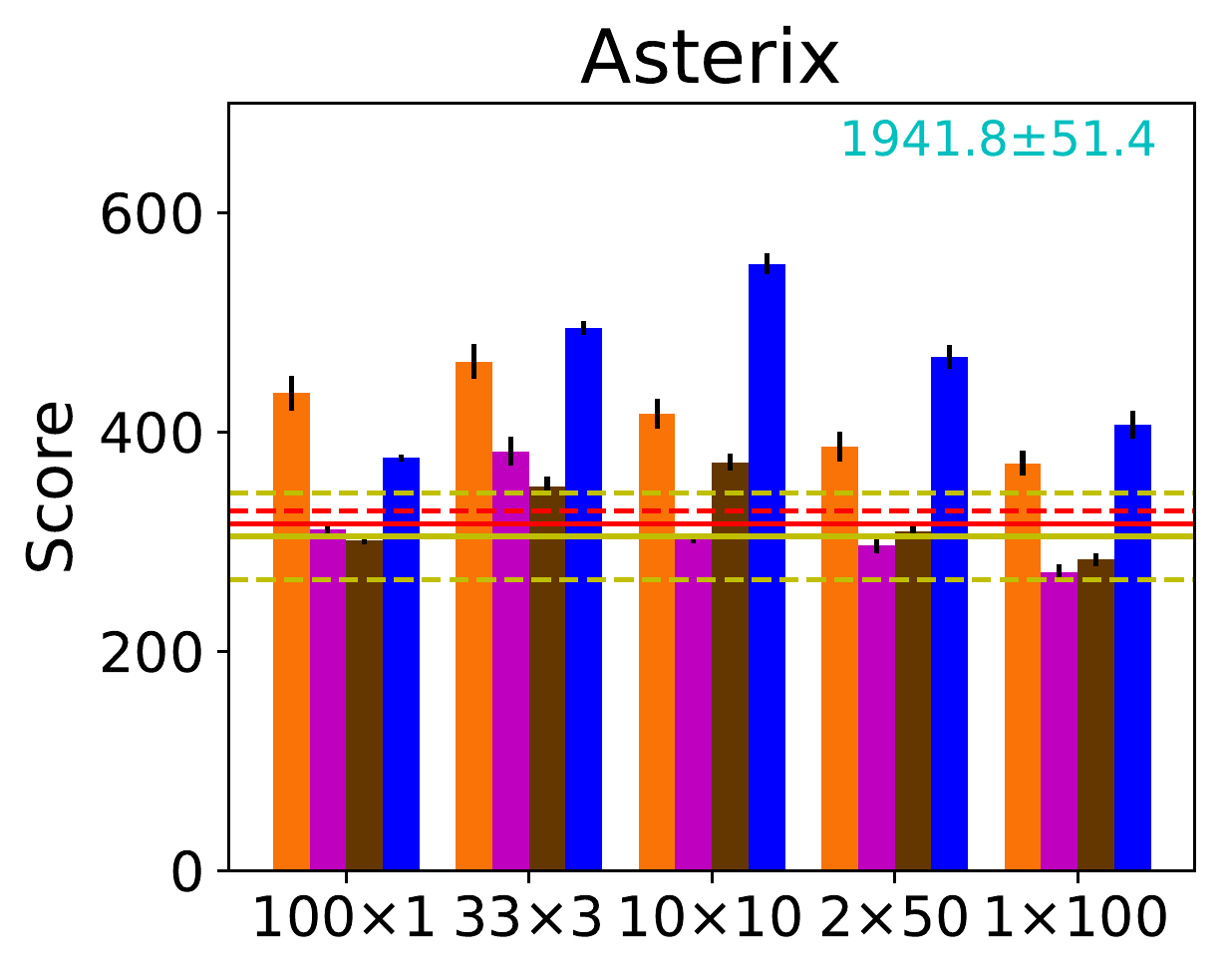}} \hfill
	\subfloat{\includegraphics[height=1.28in, keepaspectratio]{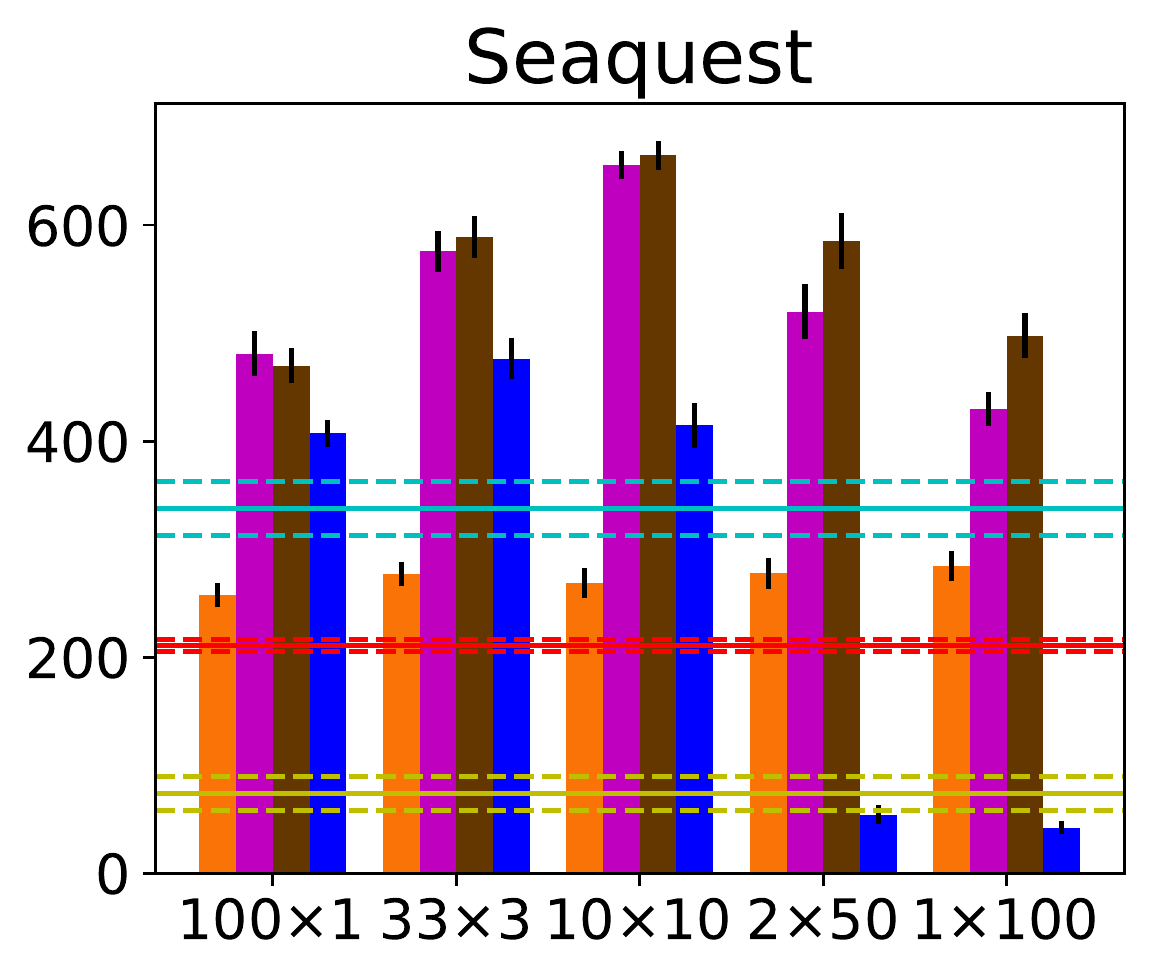}} \\
	\subfloat{\includegraphics[height=1.28in, keepaspectratio]{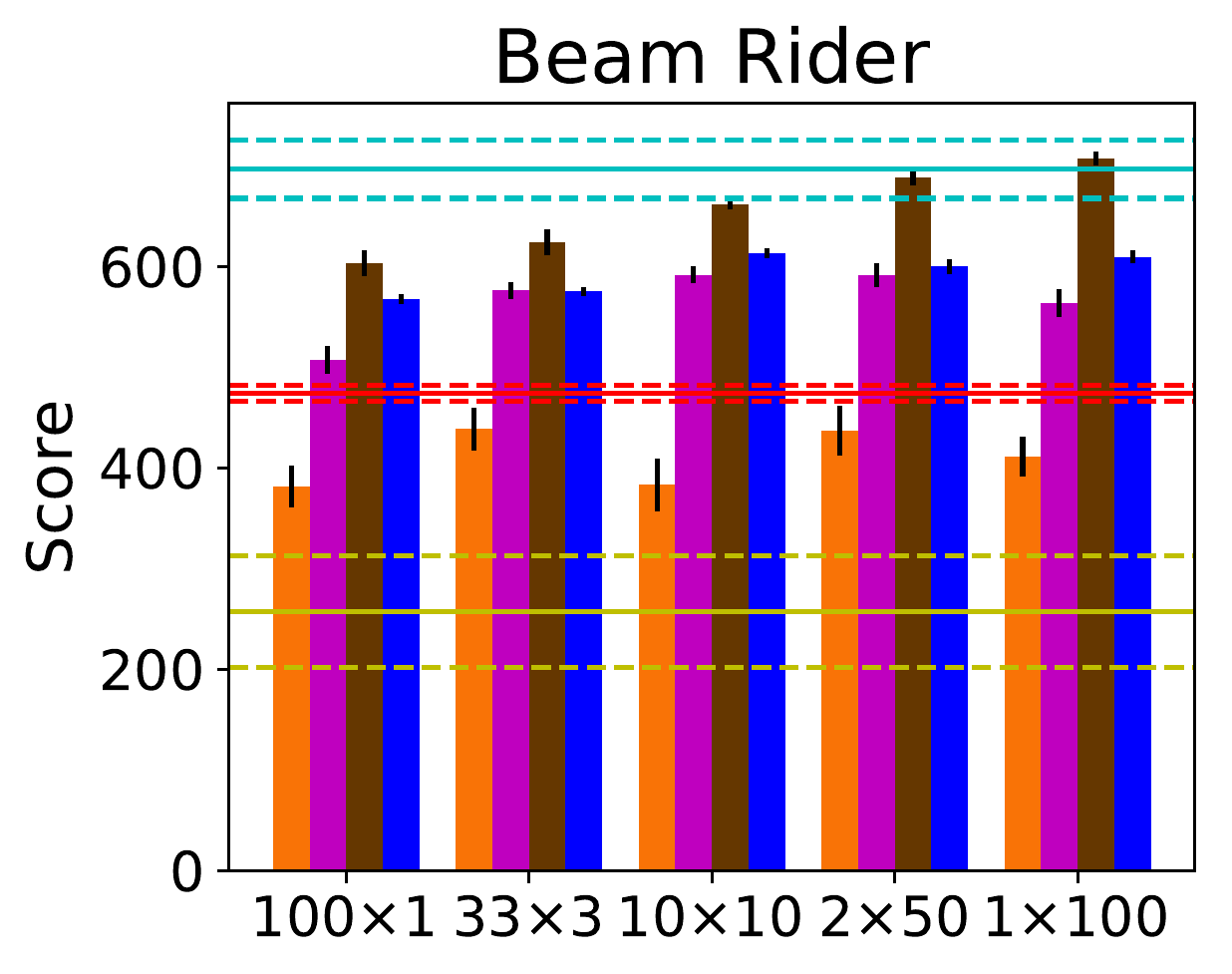}} \hfill
	\subfloat{\includegraphics[height=1.28in, keepaspectratio]{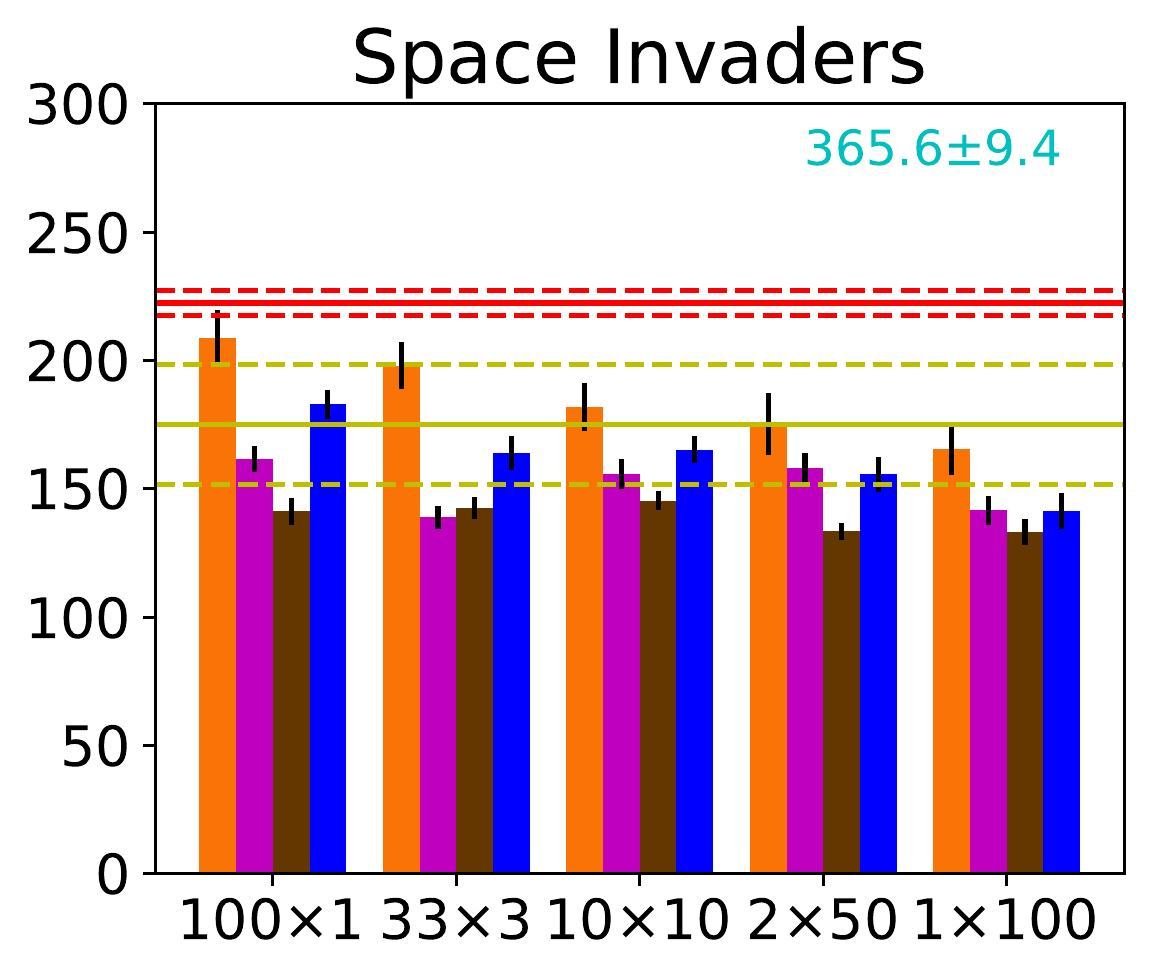}} \\
	\subfloat{\includegraphics[height=1.38in, keepaspectratio]{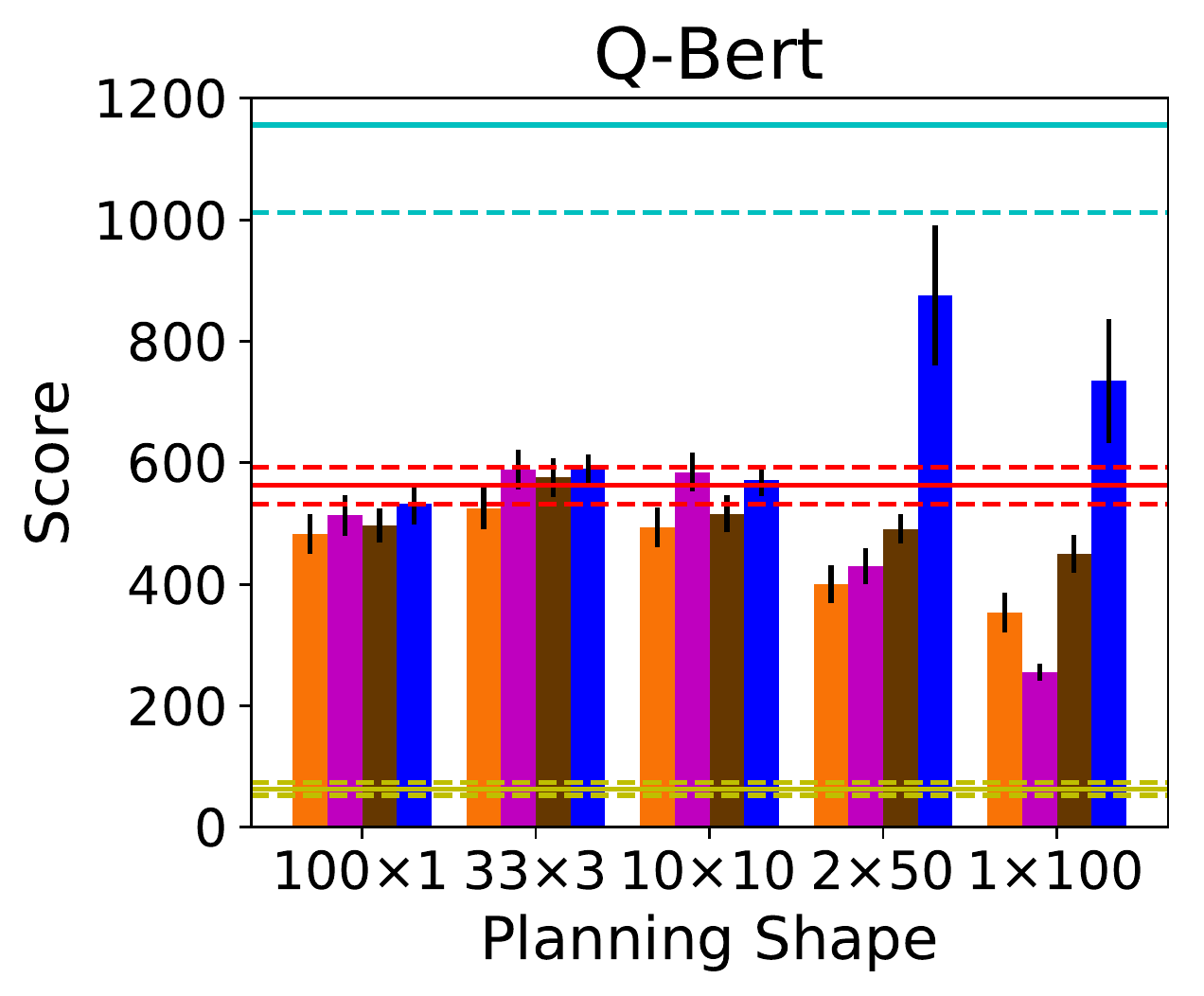}}\hfill
	\subfloat{\includegraphics[height=1.38in, keepaspectratio]{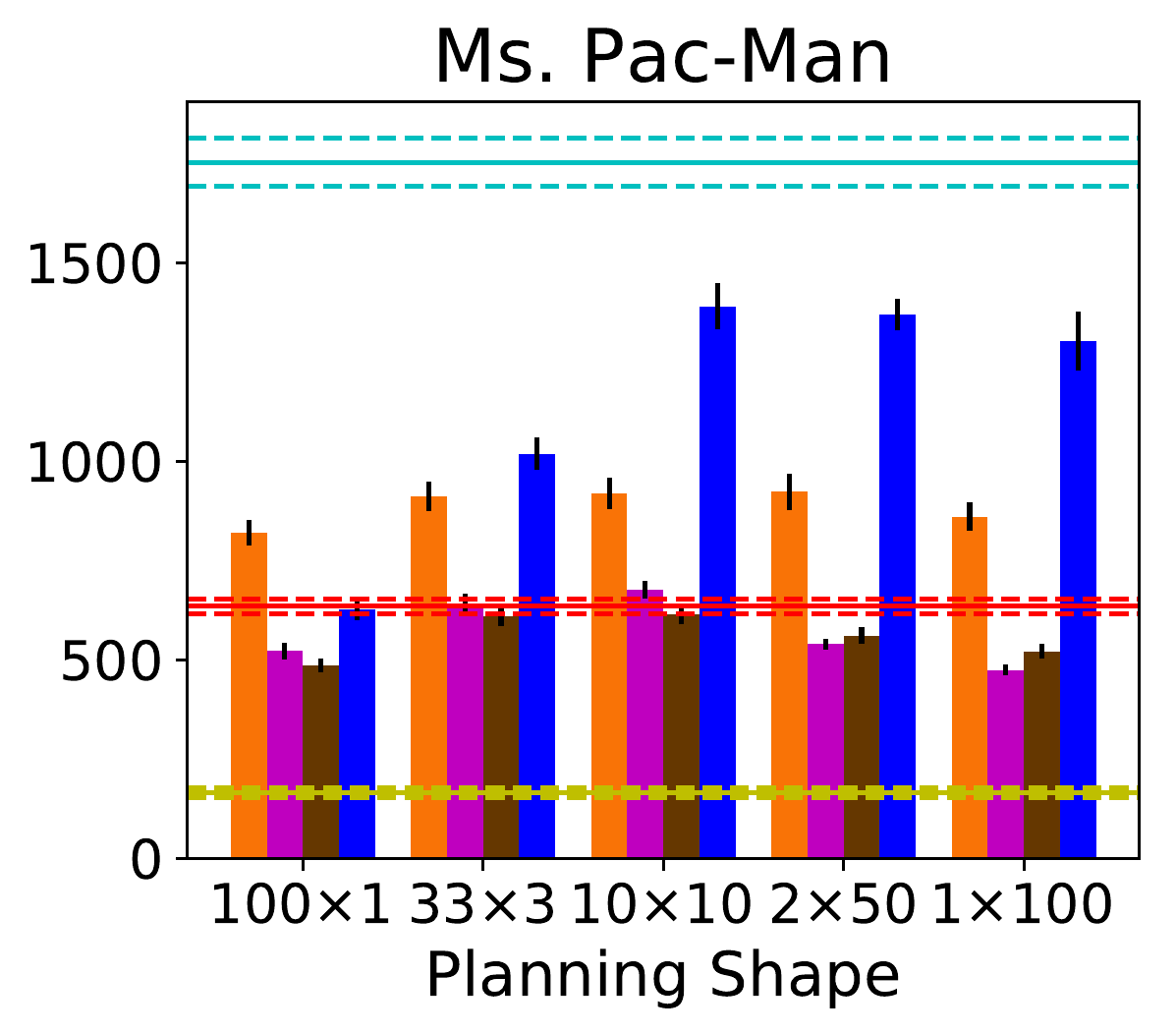}}
	\caption{Rollout-Dyna-DQN with the learned models, on six games from the ALE. 
	}
	\label{fig:allmodels}
\end{figure}

In addition to the pre-trained learned model used for the experiments presented in Section 4. We pre-trained two additional models and repeated the experiments to ensure that any trends that were observed were not specific to a particular model.
Note that because each model is pre-trained on a single dataset, our results cannot be used to draw reliable conclusions about the comparative effectiveness of the different training regimes. 
Our aim is only to study the impact of model error on Rollout-Dyna-DQN. 
As such, we refer to the models merely as Models A, B, and C; Model C is the one whose results are presented in Section 4.
The results of applying Rollout-Dyna-DQN with the three imperfect models are shown in Figure~\ref{fig:allmodels}.
The baselines are the same as in Figures 1 and 2.
The results demonstrate that the observed trends --- rollouts longer than one step
provided the most benefit and planning shape plays a significant role in performance --- are not unique to a particular instantiation of the model.

To train the models, a procedure similar to what was used by \citeauthor{oh2015action} \shortcite{oh2015action} was employed. 
In addition to the extension of the architecture to enable reward prediction, there were also two other changes from the original description.
Instead of RMSProp \cite{tieleman2012rmsprop}, the Adam optimizer was used \cite{kingma2015adam}, which \citeauthor{oh2015github} \shortcite{oh2015github} found converged more quickly.
And for preprocessing the images, instead of computing and subtracting a pixelwise mean, the mean value per channel was computed and subtracted (grayscale has one channel), following  \citeauthor{chiappa2017recurrent} \shortcite{chiappa2017recurrent}.

\textbf{Model A.} For each game, a single DQN agent was trained for 10M emulator frames.
The trained agent was then run for a series of episodes without learning, and 500k transitions (frames, actions, next frames, and rewards) were recorded to create the training set.
The model was then trained, using the training set, for 1M updates with a 1-step prediction loss (batch size 32, learning rate $1\times10^{-4}$), followed by 1M updates with a 3-step prediction loss (batch size 8, learning rate $1\times10^{-5}$), for a total of 2M updates.

\textbf{Model B.} The procedure and training data was exactly the same as for Model A, except that it was trained for an additional 1M updates using a 5-step prediction loss (batch size 8, learning rate $1\times10^{-5}$), for a total of 3M updates.

\textbf{Model C.} For this model, several independent DQN agents at different times during their learning were used to collect the training data.
For each game, five independent DQN agents were trained for 10M frames. Then, 25k transitions were recorded from evaluation episodes using a snapshot of each agent at 2.5M, 5M, 7.5M, and 10M frames during their learning.
The resulting 500k transitions were then combined to create the training set.
The model was then trained for 1M updates with a 1-step prediction loss (batch size 32, learning rate $1\times10^{-4}$), followed by 500k updates with a 3-step prediction loss (batch size 8, learning rate $1\times10^{-5}$), then finally 500k updates using a 5-step prediction loss (batch size 8, learning rate $1\times10^{-5}$), for a total of 2M updates.

\textbf{Online Learned Model.} To train the model online, batches of data are sampled from the agent's real experience in the experience replay buffer.
The model is first trained on 1-step predictions using a learning rate of $1\times10^{-4}$ for 125k updates (500k agent steps, with training occurring every 4 steps), before switching to 3-step predictions with a learning rate of $1\times10^{-5}$.
The batch size is 32 for both phases of training.

\end{document}